\documentclass[11pt,a4paper]{article}
\usepackage[hyperref]{acl2017}

\usepackage{graphicx}
\usepackage{enumerate}
\usepackage{amsmath}

\usepackage{times}
\usepackage{latexsym}
\usepackage{url}
\usepackage{xspace}
\usepackage{booktabs}
\usepackage{siunitx,caption}
\usepackage{multirow}

\usepackage{graphicx}
\usepackage[export]{adjustbox}
\usepackage{arabtex}    
\usepackage{utf8}
\setcode{utf8}

\makeatletter
\g@addto@macro{\UrlBreaks}{\UrlOrds}
\makeatother

\usepackage{color}

\aclfinalcopy

\newif\ifdraft
\drafttrue

\newcommand{\secref}[2][]{Section#1~\ref{#2}\xspace}
\newcommand{\tabref}[2][]{Table#1~\ref{#2}\xspace}

\title{SemEval-2017 Task 3: Community Question Answering}
\author{{Preslav Nakov$^1$}\hspace*{2.5mm}
        {\bf Doris Hoogeveen$^2$}\hspace*{2.5mm}
        {Llu\'{i}s M\`arquez$^1$}\hspace*{2.5mm}
        {Alessandro Moschitti$^1$}\hspace*{2mm}\\
        {\bf Hamdy Mubarak$^1$} \hspace*{2mm}
        {\bf Timothy Baldwin$^2$}\hspace*{2.5mm}
        {\bf Karin Verspoor$^2$}\hspace*{2.5mm}\\
        $^1$ALT Research Group, Qatar Computing Research Institute, HBKU\\
        $^2$The University of Melbourne\\}

\date{}

\begin{document}
\maketitle

\begin{abstractx}
We describe SemEval–2017 Task 3 on Community Question Answering. This year, we reran the four subtasks from SemEval-2016:
(A) \emph{Question--Comment Similarity}, (B) \emph{Question--Question Similarity}, (C) \emph{Question--External Comment Similarity}, and (D) \emph{Rerank the correct answers for a new question in Arabic}, providing all the data from 2015 and 2016 for training, and fresh data for testing. Additionally, we added a new subtask E in order to enable experimentation with \emph{Multi-domain Question Duplicate Detection} in a larger-scale scenario, using StackExchange subforums.
A total of 23 teams participated in the task, and submitted a total of 85 runs (36 primary and 49 contrastive) for subtasks A--D. Unfortunately, no teams participated in subtask E. A variety of approaches and features were used by the participating systems to address the different subtasks. The best systems achieved an official score (MAP) of 88.43, 47.22, 15.46, and 61.16 in subtasks A, B, C, and D, respectively. 
These scores are better than the baselines, especially for subtasks A--C.
\end{abstractx}

\section{Introduction}
%\alex{Alessandro. Done!}

Community Question Answering (CQA) on web forums such as Stack Overflow\footnote{\url{http://stackoverflow.com/}} and Qatar Living,\footnote{\url{http://www.qatarliving.com/forum}} is gaining popularity, thanks to the flexibility of forums to provide information to a user \cite{Moschitti:2016:SIGIR:workshop}. Forums are moderated only indirectly via the community, rather open, and subject to few restrictions, if any, on who can post and answer a question, or what questions can be asked. On the positive side, a user can freely ask any question and can expect a variety of answers. On the negative side, it takes efforts to go through the provided answers of varying quality and to make sense of them. It is not unusual for a popular question to have hundreds of answers, and it is very time-consuming for a user to inspect them all. 

\noindent Hence, users can benefit from automated tools to help them navigate these forums, including support for finding similar existing questions to a new question, and for identifying good answers, e.g.,~by retrieving similar questions that already provide an  answer to the new question.

Given the important role that natural language processing (NLP) plays for CQA, we have organized a challenge series to promote related research for the past three years. We have provided datasets, annotated data and we have developed robust evaluation procedures in order to establish a common ground for comparing and evaluating different approaches to CQA.

In greater detail, in SemEval-2015 Task 3 ``Answer Selection in Community Question Answering'' \cite{nakov-EtAl:2015:SemEval},\footnote{\url{http://alt.qcri.org/semeval2015/task3}} we mainly targeted conventional Question Answering (QA) tasks, i.e.,~answer selection.
In contrast, in SemEval-2016 Task 3 \cite{nakov-EtAl:2016:SemEval}, we targeted a fuller spectrum of CQA-specific tasks, moving closer to the real application needs,\footnote{A system based on SemEval-2016 Task 3 was integrated in Qatar Living's betasearch \cite{hoque-EtAl:2016:COLINGDEMO}:\\ \indent \url{http://www.qatarliving.com/betasearch}} particularly in Subtask C, which was defined as follows: ``given (\emph{i})~a new question and (\emph{ii})~a large collection of question-comment threads created by a user community, rank the comments that are most useful for answering the new question''.
%In our SemEval task, 
A test question is new with respect to the forum, but can be related to one or more questions that have been previously asked in the forum. The best answers 
%to that new question 
can come from different question--comment threads. 
%In the collection, 
The threads are independent of each other, the lists of comments are chronologically sorted, and there is meta information, e.g., date of posting, who is the user who asked/answered the question, category the question was asked in, etc.

\noindent The comments in a thread are intended to answer the question initiating that thread, but since this is a resource created by a community of casual users, there is a lot of noise and irrelevant material, in addition to the complications of informal language use, typos, and grammatical mistakes. Questions in the collection can also be related in different ways, although there is in general no explicit representation of this structure.

In addition to Subtask C, we designed subtasks A and B to give participants the tools to create a CQA system to solve subtask C. 
Specifically, Subtask A (\emph{Question-Comment Similarity}) is defined as follows: ``given a question from a question--comment thread, rank the comments according to their relevance (similarity) with respect to the question.'' Subtask B (\emph{Question-Question Similarity}) is defined as follows: ``given a new question, rerank all similar questions retrieved by a search engine, assuming that the answers to the similar questions should also answer the new question.'' 

\begin{figure}[t]
\centering
\hspace*{-4mm}
\includegraphics[width=.50\textwidth]{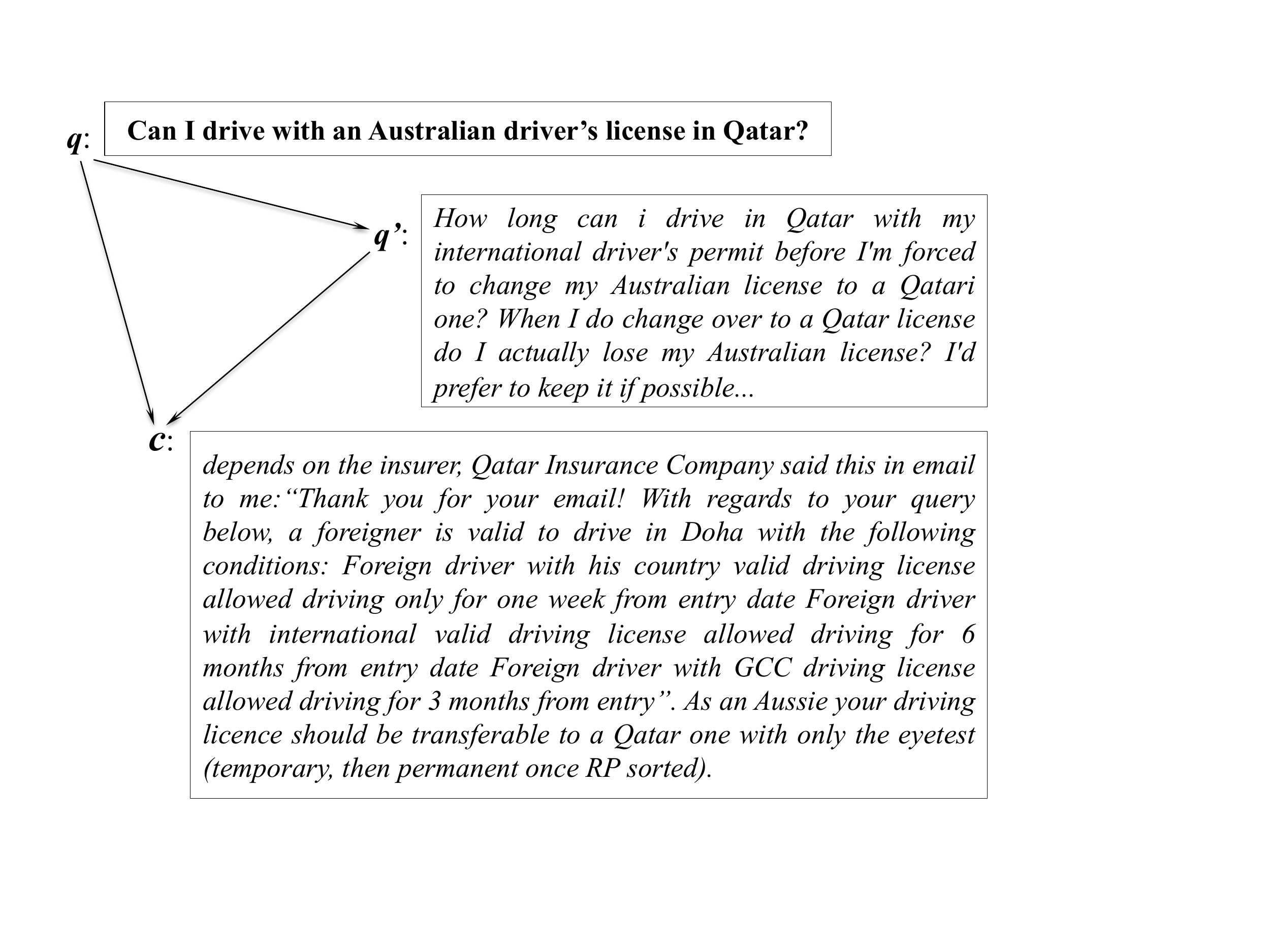}
\caption{\label{fig:triangle}The similarity triangle for CQA, showing the three pairwise interactions between the original question $q$, the related question $q'$, and a comment $c$ in the related question's thread.}
\end{figure}

The relationship between subtasks A, B, and C is illustrated in Figure~\ref{fig:triangle}. In the figure, $q$ stands for the new question, $q'$ is an existing related question, and $c$ is a comment within the thread of question $q'$. 
The edge $\overline{qc}$ relates to the main CQA task (subtask C), i.e., deciding whether a comment for a potentially related question is a good answer to the original question. This relation captures the \emph{relevance} of $c$ for $q$. 
The edge $\overline{qq'}$ represents the similarity between the original and the related questions (subtask B). This relation captures the \emph{relatedness} of $q$ and $q'$. 
Finally, the edge $\overline{q'c}$ represents the decision of whether $c$ is a good answer for the question from its thread, $q'$ (subtask A). This relation captures the \emph{appropriateness} of $c$ for $q'$. 
In this particular example, $q$ and $q'$ are indeed related, and $c$ is a good answer for both $q'$ and $q$.

The participants were free to approach Subtask C with or without solving Subtasks A and B, and participation in the main subtask and/or the two subtasks was optional. 
%A more precise definition of the different subtasks can be found in \cite{nakov-EtAl:2016:SemEval}.

We had three objectives for the first two editions of our task:
%in 2015 and 2016: 
(\emph{i})~to focus on semantic-based solutions beyond simple ``bag-of-words'' representations and ``word matching'' techniques; (\emph{ii})~to study new NLP challenges arising in the CQA scenario, e.g., relations between the comments in a thread, relations between different threads, and question-to-question similarity; and (\emph{iii}) to facilitate the participation of non-IR/QA experts. 

\noindent The third objective was achieved by 
%explicitly 
providing the set of potential answers 
%--- we did the initial retrieval step from the collection --- 
and asking the participants to (re)rank the answers, and also by defining two optional subtasks (A and B), in addition to the main subtask (i.e., C).

Last year, we were successful in attracting a large number of participants to all subtasks. However, as the task design was new (we added subtasks B and C in the 2016 edition of the task), we felt that participants would benefit from a rerun, with new test sets for subtasks A--C.
%in order to develop more customized approaches. 
%For example, last year only one team experimented with a deep learning approach, in combination with other methods. 
%Thus, we thought it would be useful to rerun the challenge, with new test sets for subtasks A--C.

We preserved the multilinguality aspect (as in 2015 and 2016), providing data for two languages: English and Arabic. In particular, we had an Arabic subtask D, which used data collected from three medical forums. This year, we used a slightly different procedure for the preparation of test set compared to the way the training, development, and test data for subtask D was collected last year. 

Additionally, we included a new subtask, subtask E, which enables experimentation on \emph{Question--Question Similarity} on a large-scale CQA dataset, i.e., StackExchange, based on the CQADupStack data set \cite{hoogeveen2015cqadupstack}.
Subtask E is a \emph{duplicate question detection} task, and like Subtask B, it is focused on question--question similarity. Participants were asked to rerank 50 candidate questions according to their relevance with respect to each query question. The subtask included several elements that differentiate it from Subtask B (see \secref{taskE}).

\noindent We provided manually annotated training data for both languages and for all subtasks. All examples were manually labeled by a community of annotators using a crowdsourcing platform. The datasets and the annotation procedure for the old data for subtasks A, B and C are described in \cite{nakov-EtAl:2016:SemEval}. In order to produce the new data for Subtask D, we used a slightly different procedure compared to 2016, which we describe in \secref{arabic-data}.
%; examples are shown on \figref[s]{GoogleArabicSearch}, \ref{AltibbiQA} and \ref{ArabicQAAnnotation}.

The remainder of this paper is organized as follows:
\secref{sec:related} introduces related work.
\secref{taskdef} gives a more detailed definition of the subtasks; it also describes the datasets and the process of their creation, and it explains the evaluation measures we used.
\secref{sec:results} presents the results for all subtasks and for all participating systems.
\secref{sec:discussion} summarizes the main approaches used by these systems and provides further discussion.
Finally, \secref{sec:conclusion} presents the main conclusions.
%that can be drawn.

%Thus, we used the fours data of last year by only changing test sets for the subtasks: (A) \emph{Question--Comment Similarity}, (B) \emph{Question--Question Similarity}, and (C)\emph{Question--External Comment Similarity} and (D) \emph{Rerank the correct answers for a new question in Arabic}, using all the past data for training and fresh data for testing. Additionally, we provided the new subtask E, to enable experimentation with \emph{Question--Question Similarity} on larger scale scenario from four StackExchange subforums: \emph{Android}, \emph{English}, \emph{Gaming}, and \emph{Wordpress}.

%%%---------------------------------------------------- R e l a t e d   W o r k
\section{Related Work}
\label{sec:related}

%\alex{Alessandro: Done! to be revised}

The first step to automatically answer questions on CQA sites is to retrieve a set of questions similar to the question that the user has asked. This set of similar questions is then used to extract possible answers for the original input question.  Despite its importance, question similarity for CQA is a hard task due to problems such as the ``lexical gap'' between the two questions. 

\emph{Question-question similarity} has been featured as a subtask (subtask B) of SemEval-2016 Task 3 on Community Question Answering \cite{nakov-EtAl:2016:SemEval}; there was also a similar subtask as part of SemEval-2016 Task 1 on Semantic Textual Similarity \cite{agirre-EtAl:2016:SemEval1}. Question-question similarity is an important problem with application to question recommendation, question duplicate detection, community question answering, and question answering in general.
Typically, it has been addressed using a variety of textual similarity measures. Some work has paid attention to modeling the question topic, which can be done explicitly, e.g., using question topic and focus \cite{duan2008searching} or using a graph of topic terms \cite{Cao:2008:RQU:1367497.1367509}, or implicitly, e.g., using a language model with a smoothing method based on the category structure of Yahoo!~Answers \cite{cao2009use} or using LDA topic language model that matches the questions not only at the term level but also at the topic level \cite{zhang2014question}.

\noindent Another important aspect is syntactic structure, e.g., \citet{wang2009syntactic} proposed a retrieval model for finding similar questions based on the similarity of syntactic trees, and \citet{DaSanMartino:CIKM:2016} used syntactic kernels. Yet another emerging approach is to use neural networks, e.g., \citet{dossantos-EtAl:2015:ACL-IJCNLP} used convolutional neural networks (CNNs), \citet{Romeo:2016coling} used long short-term memory (LSTMs) networks with neural attention to select the important part of text when comparing two questions, and \citet{LeiJBJTMM16} used a combined  recurrent--convolutional model to map questions to continuous semantic representations.
%, to mention just a few. 
Finally, translation \cite{jeon2005finding,zhou2011phrase} and cross-language models \cite{SIGIR2017:cross-lang} have also been popular for question-question similarity.

%%%%%%%%%%%%

%Different approaches have been proposed to overcome this problem. Early methods used statistical machine translation techniques to compute the semantic similarity between two questions.
%For instance, \newcite{zhou2011phrase} applied a phrase-based translation model.  Their experiments on Yahoo!~Answers showed that models based on phrases are more effective than those using words, as they are able to capture contextual information. 
%However, approaches based on SMT need a lot of data to be able to learn their parameters.

%Algorithms that try to go beyond a simple text representation are presented in \cite{cao2009use} and \cite{duan2008searching}. 
%\newcite{cao2009use} designed a similarity measure between two questions on Yahoo!~Answers, using a language model with a smoothing method based on the category structure of Yahoo!~Answers.
%\newcite{duan2008searching} searched for semantically-similar questions by identifying the question topic and focus. Specifically, they computed the similarity between the topic of a question, which represents general user interests, and the focus of a question. 

\emph{Question-answer similarity} has been a subtask (subtask A) of our task in its two previous editions \cite{nakov-EtAl:2015:SemEval,nakov-EtAl:2016:SemEval}.
This is a well-researched problem in the context of general question answering.
One research direction has been to try to match the syntactic structure of the question to that of the candidate answer. For example, \newcite{wang:2007} proposed a probabilistic quasi-synchronous grammar to learn syntactic transformations from the question to the candidate answers. \newcite{heilman:naacl:2010} used an algorithm based on Tree Edit Distance (TED) to learn tree transformations in pairs. \newcite{wang_manning:acl:2010} developed a probabilistic model to learn tree-edit operations on dependency parse trees. \newcite{yao:naacl:2013} applied linear chain conditional random fields (CRFs) with features derived from TED to learn associations between questions and candidate answers.
Moreover, syntactic structure was central for some of the top systems that participated in SemEval-2016 Task 3 \cite{SemEval2016:task3:KeLP,SemEval2016:task3:ConvKN}.

Another important research direction has been on using neural network models for question-answer similarity \cite{feng2015applying,severyn2015sigir,wang-nyberg:2015:ACL-IJCNLP, tan2015lstm,SemEval2016:task3:ConvKN,SemEval2016:task3:KeLP,SemEval2016:task3:SLS}. 
For instance, \newcite{tan2015lstm} used neural attention over a bidirectional long short-term memory (LSTM) neural network in order to generate better answer representations given the questions.
Another example is the work of \newcite{DBLP:conf/cikm/TymoshenkoBM16}, who combined neural networks with syntactic kernels.

\noindent Yet another research direction has been on using machine translation models as features for question-answer similarity
\cite{Berger:2000:BLC:345508.345576,Echihabi:2003:NAQ:1075096.1075099,jeon2005finding,Soricut:2006:AQA:1127331.1127342,riezler-EtAl:2007:ACLMain,li2011improving,Surdeanu:2011:LRA:2000517.2000520,tran-EtAl:2015:SemEval,SemEval2016:task3:UniMelb,SemEval2016:task3:ICL00},
e.g., a variation of IBM model 1 \cite{Brown:1993:MSM}, to compute the probability that the question is a ``translation'' of the candidate answer. Similarly, \cite{guzman-marquez-nakov:2016:P16-2,guzman-nakov-marquez:2016:SemEval} ported an entire machine translation evaluation framework \cite{guzman-EtAl:2015:ACL-IJCNLP} to the CQA problem.

%% LM: not well resolved: SemEval-2016 but then it is mainly other prior publications on QA
%One important aspect that emerged from SemEval-2016 \cite{nakov-EtAl:2016:SemEval} is the need for syntactic structure, which was used in all the subtasks 
%following the models proposed by 
%\cite{sigir12,severyn-moschitti:2013:EMNLP,Tymoshenko:2015:AIS:2806416.2806490,SemEval2016:task3:KeLP,SemEval2016:task3:ConvKN}.  
%% LM: jump into answer selection without telling; this sentence is too vague
%Answer selection may use models for textual entailment, semantic similarity, or for natural language inference in general. 
%In particular, we target syntactic and semantic representations of the question and answer pairs based on rich feature  representations to learn the relation: ``an answer correctly respond to a given question''.

Using information about the answer thread is another important direction, which has been explored mainly to address Subtask A.
In the 2015 edition of the task, the top participating systems used thread-level features, in addition to local features that only look at the question--answer pair.
For example, the second-best team, HITSZ-ICRC, used as a feature the position of the comment in the thread, such as whether the answer is first or last
\cite{hou-EtAl:2015:SemEval1}.
Similarly, the third-best team, QCRI, used features to model a comment in the context of the entire comment thread, focusing on user interaction~\cite{nicosia-EtAl:2015:SemEval}.
Finally, the fifth-best team, ICRC-HIT, treated the answer selection task as a sequence labeling problem and proposed recurrent convolutional neural networks to recognize good comments \cite{zhou-EtAl:2015:SemEval}.

In follow-up work, \newcite{zhou-EtAl:2015:ACL-IJCNLP} included long-short term memory (LSTM) units in their convolutional neural network to model the classification sequence for the thread, and \newcite{barroncedeno-EtAl:2015:ACL-IJCNLP} exploited the dependencies between the thread comments to tackle the same task. This was done by designing features that look globally at the thread and by applying structured prediction models, such as CRFs.
%conditional random fields \cite{Lafferty01}. 

%%LM: I reduced this section; joining the two paragraphs and reducing the second paper by Joty et al.
This research direction was further extended by \newcite{joty:2015:EMNLP}, who used the output structure at the thread level in order to make more consistent global decisions about the goodness of the answers in the thread. They modeled the relations between pairs of comments at any distance in the thread, and combined the predictions of local classifiers using graph-cut and Integer Linear Programming. %(ILP). 
In follow up work, 
%Finally, 
\newcite{Joty:2016:NAACL} proposed joint learning models that integrate inference within the learning process using global normalization and an Ising-like edge potential.
%One of these models, an instance of a fully connected pairwise CRF, yielded the best results on the task (SemEval-2015 Task 3) to date. 
%The first one jointly learns two node- and edge-level maxent classifiers with stochastic gradient descent and integrates the inference step with loopy belief propagation. 
%The second model is an instance of fully connected pairwise CRFs (FCCRF). The FCCRF model significantly outperforms all other approaches and yields the best results on the task (SemEval-2015 Task 3) to date. %Crucial elements for its success are the global normalization and an Ising-like edge potential.

%%% LM: This paragraph is totally disconnected and contains only self citations
%\newcite{Tymoshenko:NAACL:2016} combines tree kernels and neural networks for question answering whereas in
%\newcite{DBLP:conf/cikm/TymoshenkoBM16}, they produced an advanced system using tree kernels and neural networks for Subtask A of SemEval CQA.  
%\newcite{DBLP:conf/cikm/MartinoBR0M16} built an advanced system for Subtask B using kernels over trees and rank features. In particular, the authors introduced an effective text selection technique for improving the speed and effectiveness of the tree kernel approach \cite{RomeoMBMBHZMG16,DBLP:conf/coling/Barron-CedenoMR16,Romeo2017}.

%%% LM: I have rewritten the contribution of the first paper and eliminated the arXiv citation

\emph{Question--External comment similarity} is our main task (subtask C), and it is inter-related to subtasks A and B, as described in the triangle of Figure~\ref{fig:triangle}. This task has been much less studied in the literature, mainly because its definition is specific to our SemEval Task 3, and it first appeared in the 2016 edition~\cite{nakov-EtAl:2016:SemEval}.
Most of the systems that took part in the competition, including the winning system of the SUper team \cite{SemEval2016:task3:SUper}, approached the task indirectly by solving subtask A at the thread level and then using these predictions together with the reciprocal rank of the related questions in order to produce a final ranking for subtask C.
One exception is the \emph{KeLP} system \cite{SemEval2016:task3:KeLP}, which was ranked second in the competition. This system combined information from different subtasks and from all input components. It used a modular kernel function, including stacking from independent subtask A and B classifiers, and applying SVMs to train a Good vs. Bad classifier \cite{SemEval2016:task3:KeLP}. 
In a related study, \newcite{nakov-marquez-guzman:2016:EMNLP2016} discussed the input information to solve Subtask C, and concluded that one has to model mainly question-to-question similarity (Subtask B) and answer goodness (subtask A), while modeling the direct relation between the new question and the candidate answer (from a related question) was found to be far less important.  

Finally, in another recent approach, \newcite{bonadiman-uva-moschitti:2017:EACLshort} studied how to combine the different CQA subtasks. They presented a multitask neural architecture where the three tasks are trained together with the same representation. The authors showed that the multitask system yields good improvement for Subtask C, which is more complex and clearly dependent on the other two tasks.

\emph{Some notable features across all subtasks.} Finally, we should mention some interesting features used by the participating systems across all three subtasks. This includes fine-tuned word embeddings\footnote{\url{https://github.com/tbmihailov/semeval2016-task3-cqa}} \cite{SemEval2016:task3:SemanticZ}; features modeling text complexity, veracity, and user trollness\footnote{Using a heuristic that if several users call somebody a troll, then s/he should be one \cite{mihaylov-georgiev-nakov:2015:CoNLL,mihaylov-EtAl:2015:RANLP2015,mihaylov-nakov:2016:trolls,DarkWeb:2017}.}
\cite{SemEval2016:task3:SUper}; sentiment polarity features \cite{nicosia-EtAl:2015:SemEval}; and PMI-based goodness polarity lexicons \cite{SemEval2016:task3:PMI-cool,PMI:SIGIR:2017}. 

\begin{table*}[t]
\small
\begin{center}
\begin{tabular}{l@{}rrrr}
\multirow{2}{*}{\bf Category} & \bf Train+Dev+Test & \bf Train(1,2)+Dev+Test & \multirow{2}{*}{\bf Test}\\
 & \bf from SemEval-2015 & \bf from SemEval-2016 & \\
\hline
\hline
\bf Original Questions & \multicolumn{1}{c}{\bf --} & \bf (200+67)+50+70 & \bf 88 & \\
\\
\bf Related Questions & \bf 2,480+291+319 & \bf (1,999+670)+500+700 & \bf 880  \\
 -- Perfect Match & \multicolumn{1}{c}{--} &  (181+54)+59+81 &  24 \\
 -- Relevant & \multicolumn{1}{c}{--} &  (606+242)+155+152 &  139 \\
 -- Irrelevant & \multicolumn{1}{c}{--}  &   (1,212+374)+286+467 &  717 \\
  & \bf  & \bf  & \bf \\
\bf Related Comments & \multicolumn{1}{c}{\bf --} & \bf (19,990+6,700)+5,000+7,000 & \bf 8,800\\
\bf (with respect to Original Question)  & &   &   &  \\
 -- Good & \multicolumn{1}{c}{--} &  (1,988+849)+345+654 &    246\\
 -- Bad  & \multicolumn{1}{c}{--} &  (16,319+5,154)+4,061+5,943  &   8,291 \\
 -- Potentially Useful & \multicolumn{1}{c}{--} &  (1,683+697)+594+403 &   263 \\
 & \bf  & \bf  & \bf \\
\bf Related Comments & \bf 14,893+1,529+1,876 & \bf (14,110+3,790)+2,440+3,270 & \bf 2,930\\
\bf (with respect to Related Question)  &   & &   &  \\
 -- Good &  7,418+813+946 &  (5,287+1,364)+818+1,329 &   1,523 \\
 -- Bad  &  5,971+544+774 &  (6,362+1,777)+1,209+1,485 &   1,407 \\
 -- Potentially Useful &  1,504+172+156 &  (2,461+649)+413+456 &   0 \\
\hline
\end{tabular}
\caption{Statistics about the English CQA-QL dataset. 
		 Note that the \emph{Potentially Useful} class was merged with \emph{Bad} at test time for SemEval-2016 Task 3, and was eliminated altogether at SemEval-2017 task 3.}
\label{table:statistics:english}
\end{center}
\end{table*}

\section{Subtasks and Data Description}
\label{taskdef}
\label{sec:task}

The 2017 challenge was structured as a set of five subtasks, four of which (A, B, C and E) were offered for English, while the fifth (D) one was for Arabic. 
%We describe them below in detail. 
We leveraged the data we developed in 2016 for the first four subtasks, creating only new test sets for them, whereas we built a completely new dataset for the new Subtask E.

\subsection{Old Subtasks}
The first four tasks and the datasets for them are described in \cite{nakov-EtAl:2016:SemEval}. Here we review them briefly.
%, and we define the new Subtask E.
%
\paragraph{English subtask A} \emph{Question-Comment Similarity}.
Given a question $Q$ and the first ten comments\footnote{We limit the number of comments we consider to the first ten only in order to spare some annotation efforts.} in its question thread ($c_1,\dots,c_{10}$), the goal is to rank these ten comments according to their relevance with respect to that question.

Note that this is a ranking task, not a classification task; we use mean average precision (MAP) as an official evaluation measure. This setting was adopted as it is closer to the application scenario than pure comment classification. For a perfect ranking, a system has to place all ``Good'' comments above the ``PotentiallyUseful'' and the ``Bad'' comments; the latter two are not actually distinguished and are considered ``Bad'' at evaluation time. This year, we elliminated the ``PotentiallyUseful'' class for test at annotation time.

%Note also that subtask A this year is the same as subtask A at SemEval-2015 Task 3, but with slightly different annotation and evaluation measure.

\paragraph{English subtask B} \emph{Question-Question Similarity}.
Given a new question $Q$ (aka \emph{original question}) and the set of the first ten related questions from the forum ($Q_1,\dots,Q_{10}$) retrieved by a search engine, the goal is to rank the related questions according to their similarity with respect to the original question. 

In this case, we consider the ``PerfectMatch'' and the ``Relevant'' questions both as good (i.e., we do not distinguish between them and we will consider them both ``Relevant''), and they should be ranked above the ``Irrelevant'' questions. 
As in subtask A, we use MAP as the official evaluation measure. To produce the ranking of related questions, participants have access to the corresponding related question-thread.\footnote{Note that the search engine indexes entire Web pages, and thus, the search engine has compared the original question to the related questions together with their comment threads.} 
Thus, being more precise, this subtask could have been named \emph{Question --- Question+Thread Similarity}.

\paragraph{English subtask C} \emph{Question-External Comment Similarity}.
Given a new question $Q$ (also known as the \emph{original question}), and the set of the first ten related questions ($Q_1,\dots,Q_{10}$) from the forum retrieved by a search engine for $Q$, each associated with its first ten comments appearing in $Q$'s thread ($c_1^1,\dots,c_1^{10},\dots,c_{10}^1,\dots,c_{10}^{10}$), the goal is to rank these 10$\times$10 = 100 comments $\{c_i^j\}_{i,j=1}^{10}$ according to their relevance with respect to the original question $Q$. 

\noindent This is the main English subtask.
As for subtask A, we want the ``Good'' comments to be ranked above the ``PotentiallyUseful'' and the ``Bad'' comments, which will be considered just bad in terms of evaluation. Although, the systems are supposed to work on 100 comments, we take an application-oriented view in the evaluation,
assuming that
%: we assume that  potential users are presented with a relatively short list of candidate answers (e.g., ten, as is common for search engines today). Thus, the 
users would like to have good comments concentrated in the first ten positions.
%, (i.e., all good comments should be ranked before any non-good comment). 
We believe users care much less about what happens in lower positions (e.g., after the 10th) in the rank, as they typically do not ask for the next page of results in a search engine such as Google or Bing. This is reflected in our primary evaluation score, MAP, which we restrict to consider only the top ten results for subtask C. 

\paragraph{Arabic subtask D} \emph{Rank the correct answers for a new question}.
Given a new question $Q$ (aka the original question), the set of the first 30 related questions retrieved by a search engine, each associated with one correct answer ($(Q_1,c_1)\dots,(Q_{30},c_{30})$), the goal is to rank the 30 question-answer pairs according to their relevance with respect to the original question. We want the ``Direct'' and the ``Relevant'' answers to be ranked above the ``Irrelevant'' answers; the former two are considered ``Relevant'' in terms of evaluation. We evaluate the position of ``Relevant'' answers in the rank, and this is again a ranking task.
Unlike the English subtasks, here we use 30 answers since the retrieval task is much more difficult, leading to low recall, and the number of correct answers is much lower. Again, the systems were evaluated using MAP, restricted to the top-10 results.

\subsubsection{Data Description for A--D}

The English data for subtasks A, B, and C comes from the Qatar Living forum, which is organized as a set of seemingly independent question--comment threads.  In short, for subtask A, we annotated the comments in a question-thread as ``Good'', ``PotentiallyUseful'' or ``Bad'' with respect to the question that started the thread. Additionally, given original questions, we retrieved related question--comment threads and annotated the related questions as ``PerfectMatch'', ``Relevant'', or ``Irrelevant'' with respect to the original question (Subtask B). We then annotated the comments in the threads of related questions as ``Good'', ``PotentiallyUseful'' or ``Bad'' with respect to the original question (Subtask C).

\noindent For Arabic, the data was extracted from medical forums and has a different format. Given an original question, we retrieved pairs of the form (related\_question, answer\_to\_the\_related\_question). These pairs were annotated as ``Direct'' answer, ``Relevant'' and ``Irrelevant'' with respect to the original question.

%\subsubsection{New Test Data for Subtasks A, B, and C}

\paragraph{For subtasks A, B, and C} we annotated new English test data following the same setup as for SemEval-2016 Task 3 \cite{nakov-EtAl:2016:SemEval}, except that we elliminated the ``Potentially Useful'' class for subtask A. We first selected a set of questions to serve as original questions. In a real-world scenario those would be questions that had never been asked previously, but here we used existing questions %already existing in 
from Qatar Living. 

From each original question, we generated a query, using the question's subject (after some word removal if the subject was too long). Then, we executed the query against Google, limiting the search to the Qatar Living forum, and we collected up to 200 resulting question-comment threads as related questions. Afterwards, we filtered out threads with less than ten comments as well as those for which the question was more than 2,000 characters long. Finally, we kept the top-10 surviving threads, keeping just the first 10 comments in each thread.

We formatted the results in XML with UTF-8 encoding, adding metadata for the related questions and for their comments; however, we did not provide any meta information about the original question, in order to emulate a scenario where it is a new question, never asked before in the forum. In order to have a valid XML, we had to do some cleansing and normalization of the data. We added an XML format definition at the beginning of the XML file and we made sure it validated.

We organized the XML data as a sequence of original questions (OrgQuestion), where each question has a subject, a body, and a unique question identifier (ORGQ\_ID). Each such original question is followed by ten threads, where each thread consists of a related question (from the search engine results) and its first ten comments.

We made available to the participants for training and development the data from 2016 (and for subtask A, also from 2015), and we created a new test set of 88 new questions associated with 880 question candidates and 8,800 comments; details are shown in Table~\ref{table:statistics:english}.

%\alex{Hamdy/Preslav. Description of the annotation process for the new test sets. The idea is to heavily refer to \cite{nakov-EtAl:2016:SemEval} and simply telling that we annotated other data. Maybe we change a bit the crowdflower task?}

\begin{table}[t]
\small
\begin{center}
\begin{tabular}{lrrrr}
\multirow{2}{*}{\bf Category} & \multicolumn{3}{c}{\bf SemEval-2016 data} & \multirow{2}{*}{\bf Test-2017}\\
\cline{2-4}
 & \bf Train & \bf Dev & \bf Test & \\
\hline
\hline
\bf Questions & \bf 1,031 & \bf 250 & \bf 250 & \bf 1,400\\
\bf QA Pairs &  \bf 30,411 &  \bf 7,384 & \bf 7,369 & \bf 12,600\\
 -- Direct &  917 &  70 &  65 &  891\\
 -- Related &  17,412 &  1,446 &  1,353 &  4,054\\
 -- Irrelevant &  12,082 &  5,868 &  5,951 &  7,655\\
\hline
\end{tabular}
\caption{Statistics about the CQA-MD corpus.}
\label{table:statistics:arabic}
\end{center}
\end{table}

%\subsubsection{Data Description for D}
\label{arabic-data}
%\alex{Hamdy. Fixing the description below. Here we did a different annotation task but I see that it is taken into account.}

%We created a new Arabic test dataset, following a slightly different procedure compared to 2016.

%\paragraph{Data Collection}

\paragraph{For subtasks D} we had to annotate new test data.
In 2016, we used data from three Arabic medical websites, which we downloaded and indexed locally using Solr.\footnote{\url{https://lucene.apache.org/solr/}} Then, we performed 21 different query/document formulations, and we merged the retrieved results, ranking them according to the reciprocal rank fusion algorithm~\cite{cormack2009reciprocal}. Finally, we truncated the result list to the 30 top-ranked question--answer pairs.

This year we only used one of these websites, namely \url{Altibbi.com}\footnote{\url{http://www.altibbi.com/}\<طبية>-\<اسئلة>}
First, we selected some questions from that website to be used as original questions, and then we used Google to retrieve potentially related questions using the \url{site:*} filter.

%We crawled all questions from the ``Medical-Questions'' section of the \url{Altibbi.com} medical site,\footnote{http://www.altibbi.com/\<طبية>-\<اسئلة>} using the Google API\footnote{\url{http://www.google.com/search?q="QUERY"}} and the \url{site:*} filter.

We turned the question into a query as follows: We first queried Google using the first thirty words from the original question. If this did not return ten results, we reduced the query to the first ten non-stopwords\footnote{We used the following Arabic stopword list: \url{https://sites.google.com/site/kevinbouge/stopwords-lists}} from the question, and if needed we further tried using the first five non-stopwords only. If we did not manage to obtain ten results, we discarded that original question. 

If we managed to obtain ten results, we followed the resulting links and we parsed the target page 
%(an example is shown on Figure~\ref{AltibbiQA}) 
to extract the question and the answer, which is given by a physician, as well as some metadata such as date, question classification, doctor's name and country, etc.

In many cases, Google returned our original question as one of the search results, in which case we had to exclude it, thus reducing the results to nine. In the remaining cases, we excluded the 10th result in order to have the same number of candidate question--answer pairs for each original question, namely nine.
Overall, we collected 1,400 original questions, with exactly nine potentially related question--answer pairs for each of them, i.e., a total of 12,600 pairs.

%\subsubsection{Data Annotation}

\noindent We created an annotation job on CrowdFlower to obtain judgments about the relevance of the question--answer pairs with respect to the original question.
%Figure~\figref{ArabicQAAnnotation} shows a snapshot of the annotation task. 
We controlled the quality of annotation using a hidden set of 50 test questions. We had three judgments per example, which we combined using the CrowdFlower mechanism. 
The average agreement was 81\%. Table~\ref{table:statistics:arabic} shows statistics about the resulting dataset, together with statistics about the datasets from 2016, which could be used for training and development.
%, and Figure~\ref{SemEvalArabicXML} .

% Annotators were asked to judge the relation between each pair (RelQ/RelA) and OrgQ to one of the following classes: Direct answer, Useful answer, and NotUseful answer. To ensure quality, we utilized 40 test questions (40 * 9 = 360 judgments) and used 70\% as an acceptance threshold. 
%Average agreement for test questions were 80.55\% indicating high quality annotations. Distribution of annotation is: Direct = 7.07\%, Useful = 32.17\%, and NotUseful = 60.75\%.

% \begin{figure}[t]
% \centering
%   \includegraphics[width=0.48\textwidth, frame]{ArabicQAAnnotation2.png}
%   \caption{Example screen for the Arabic manual annotation task in CrowdFlower.}
%   \label{ArabicQAAnnotation}
% \end{figure}

%Figure \ref{SemEvalArabicXML} shows part of the generated XML file for Arabic CQA task. For OrgQ, we save Question ID ``QID", and Question Text ``Qtext". For RelQ/RelA pair, we save Pair ID ``QAID", RelQ Text ``QAquestion", RelA Text ``QAanswer", and annotation judgment for relation between RelQ/RelA pair and OrgQ ``QArel''.

% \begin{figure}[t]
% \centering
%   \includegraphics[width=0.48\textwidth, frame]{SemEvalArabicXML2.png}
%   \caption{Generated XML File for Arabic CQA. The literal English translations are provided for illustration, and are not part of the XML we distributed to the participants.}
%   \label{SemEvalArabicXML}
% \end{figure}

\subsubsection{Evaluation Measures for A--D}
\label{sec:scoring}
%\lluis{...I propose to have the description of the scoring metrics as a subsection of the experiments and results section (to be written by Alessandro/Preslav)}
The official evaluation measure %with respect to which the participating systems are ranked
we used to rank the participating systems is Mean Average Precision (``MAP''), calculated over the top-10 comments as ranked by a participating system. We further report the results for two unofficial ranking measures, which we also calculated over the top-10 results only: Mean Reciprocal Rank (``MRR'') and Average Recall (``AvgRec'').
Additionally, we report the results for four standard classification measures, %which are not about ranking and 
which we calculate over the full list of results: Precision, Recall and F$_1$ (with respect to the Good/Relevant class), and Accuracy.

We released a specialized scorer that calculates and returns all the above-mentioned scores.
%, among with some other.

\subsection{The New Subtask E}
\label{taskE}

Subtask E is a duplicate question detection task, similar to Subtask B. 
Participants were asked to rerank 50 candidate questions according to their relevance with respect to each query question. The subtask included several elements that distinguish it from Subtask B:

\begin{itemize}
  \setlength\itemsep{1em}
	\item Several meta-data fields were added, including the tags that are associated with each question, the number of times a question has been viewed, and the score of each question, answer and comment (the number of upvotes it has received from the community, minus the number of downvotes), as well as user statistics, containing information such as user reputation and user badges.\footnote{The complete list of available meta-data fields can be found on the Task website.}
    \item At test time, two extra test sets containing data from two surprise subforums were provided, to test the participants' system's cross-domain performance.
    \item The participants were asked to truncate their result list in such a way that only ``PerfectMatch'' questions appeared in it. The evaluation metrics were adjusted to be able to handle empty result lists (see \secref{sec:e-evaluation}).
    \item The data was taken from StackExchange instead of the Qatar Living forums, and reflected the real-world distribution of duplicate questions in having many query questions with zero relevant results.
\end{itemize}

The cross-domain aspect was of particular interest, as it has not received much attention in earlier duplicate question detection research.

\begin{table}[t]
  \centering
    \begin{tabular}{lccc}
      \toprule
      \bf Subforums     & \bf Train		& \bf Development & \bf Test 	\\
      \midrule	
      Android		& 10,360		& 3,197		& 3,531		\\
      English 		& 20,701		& 6,596		& 6,383		\\
      Gaming 		& 14,951		& 4,964		& 4,675		\\
      Wordpress 	& 13,733		& 5,007		& 3,816		\\
      \midrule
      Surprise 1	& ---		& ---		& 5,123		\\
      Surprise 2	& ---		& ---		& 4,039		\\
      \bottomrule
    \end{tabular}
  \caption{Statistics on the data for Subtask E. Shown is the number of query questions; for each of them, 50 candidate questions were provided.}
  \label{tab:stackexchange}
\end{table}

\subsubsection{Data Description for E}

The data consisted of questions from the following four StackExchange subforums: \emph{Android}, \emph{English}, \emph{Gaming}, and \emph{Wordpress}, derived from a data set known as CQADupStack \cite{hoogeveen2015cqadupstack}. Data size statistics can be found in \tabref{tab:stackexchange}. These subforums were chosen due to their size, and to reflect a variety of domains.

% DH: Shall I mention which subforums the surprise data is taken from?
% no let us keep the surprise :) for another time :)
% TJB: :-) fair enough

The data was provided in the same format as for the other subtasks. Each original question had 50 candidate questions, and these related questions each had a number of comments. On top of that, they had a number of answers, and each answer potentially had individual comments. The difference between answers and comments is that answers should contain a well-formed answer to the question, while comments contain things such as requests for clarification, remarks, and small additions to someone else's answer. Since the content of StackExchange is provided by the community, the precise delineation between comments and the main body of a post can vary across forums.

\noindent The relevance labels in the development and in the training data were sourced directly from the users of the StackExchange sites, who can vote for questions to be closed as duplicates: these are the questions we labeled as \emph{PerfectMatch}. 

The questions labeled as \emph{Related} are questions that are not duplicates, but that are somehow similar to the original question, also as judged by the StackExchange community. It is possible that some duplicate labels are missing, due to the voluntary nature of the duplicate labeling on StackExchange. The development and training data should therefore be considered a silver standard \cite{Hoogeveen+:2016a}.  

For the test data, we started an annotation project together with StackExchange.\footnote{A post made by StackExchange about the project can be found here: {\small\url{http://meta.stackexchange.com/questions/286329/project-reduplication-of-deduplication-has-begun}}} The goal was to obtain multiple annotations per question pair in the test set, from the same community that provided the labels in the development and in the training data. We expected the community to react enthusiastically, because the data would be used to build systems that can improve duplicate question detection on the site, ultimately saving the users manual effort. Unfortunately, only a handful of people were willing to annotate a sizeable set of question pairs, thus making their annotations unusable for the purpose of this shared task.

An example that includes a query question from the English subforum, a duplicate of that question, and a non-duplicate question (with respect to the query) is shown below:
\begin{itemize}
	\item Query: \textit{Why do bread companies add sugar to bread?}
	\item Duplicate: \textit{What is the purpose of sugar in baking plain bread?}
	\item Non-duplicate: \textit{Is it safe to eat potatoes that have sprouted?}
\end{itemize}

\subsubsection{Evaluation Measure for E}
\label{sec:e-evaluation}

In CQA archives, the majority of new questions do not have a duplicate in the archive. We maintained this characteristic in the training, in the development, and in the test data, to stay as close to a real world setting as possible. This means that for most query questions, the correct result is an empty list. 

\noindent This has two consequences: (1) a system that always returns an empty list is a challenging baseline to beat, and (2) standard IR evaluation metrics like MAP, which is used in the other subtasks, cannot be used, because they break down when the result list is empty or there are no relevant documents for a given query.

To solve this problem we used a modified version of MAP, as proposed by \newcite{liu2016}. To make sure standard IR evaluation metrics do not break down on empty result list queries, \newcite{liu2016} add a nominal terminal document to the end of the ranking returned by a system, to indicate where the number of relevant documents ended. This terminal document has a corresponding gain value of:
\begin{eqnarray*}
  r_{t} = \begin{cases}
    1 & \textup{if}\: R = 0 \\
    \sum_{i=1}^{d} r_{i}/R & \textup{if}\: R > 0 
  \end{cases}
\end{eqnarray*}
The result of this adjustment is that queries without relevant documents in the index, receive a MAP score of 1.0 for an empty result ranking. This is desired, because in such cases, the empty ranking is the correct result.

%%%----------------- P a r t i c i p a n t s   a n d   R e s u l t s

\section{Participants and Results}
\label{sec:results}

The list of all participating teams can be found in \tabref{table:teams}. The results for subtasks A, B, C, and D are shown in tables \ref{table:results:subtaskA}, \ref{table:results:subtaskB}, \ref{table:results:subtaskC}, and \ref{table:results:subtaskD}, respectively. Unfortunately, there were no official participants in Subtask E, and thus we present baseline results in \tabref{tab:baseline-e}. In all tables, the systems are ranked by the official MAP scores for their primary runs\footnote{Participants could submit one primary run, to be used for the official ranking, and up to two contrastive runs, which are scored, but they have unofficial status.} (shown in the third column). The following columns show the scores based on the other six unofficial measures; the ranking with respect to these additional measures are marked with a subindex (for the primary runs).

Twenty two teams participated in the challenge presenting a variety of approaches and features to address the different subtasks. They submitted a total of 85 runs (36 primary and 49 contrastive), which breaks down by subtask as follows: The English subtasks A, B and C attracted 14, 13, and 6 systems and 31, 34 and 14 runs, respectively. The Arabic subtask D got 3 systems and 6 runs. And there were no participants for subtask E.

\noindent The best MAP scores had large variability depending on the subtask, going from 15.46 (best result for subtask C) to 88.43 (best result for subtask A). The best systems for subtasks A, B, and C were able to beat the baselines we provided by sizeable margins. In subtask D, only the best system was above the IR baseline. 

 \begin{table*}[tbh]
 \small
 \begin{center}
 \begin{tabular}{@{}l@{ }@{ }l@{}}
\toprule
 \bf Team ID & \bf Team Affiliation\\
 \midrule
 Beihang-MSRA & Beihang University, Beijing, China; Microsoft Research, Beijing, China\\
 & \cite{SemEval-2017:task3:BEIHANG-MSRA}\\
 bunji & Hitachi Ltd., Japan\\
 & \cite{SemEval-2017:task3:BUNJI}\\
 ECNU & East China Normal University, P.R. China;
 Shanghai Key Laboratory of Multidimensional \\
 & Information Processing, P.R. China \cite{SemEval-2017:task3:ECNU}\\
 EICA & East China Normal University, Shanghai, P.R.China\\
 & \cite{SemEval-2017:task3:EICA}\\
 FuRongWang & National University of Defense Technology, P.R. China\\
 & \cite{SemEval-2017:task3:FURONGWANG}\\
 FA3L & University of Pisa, Italy \\
 & \cite{SemEval-2017:task3:FA3L}\\
 GW\_QA & The George Washington University, D.C. USA\\
 & \cite{SemEval-2017:task3:GW-QA}\\
 IIT-UHH & Indian Institute of Technology Patna, India; University of Hamburg, Germany\\
 & \cite{SemEval-2017:task3:IIT-UHH}\\
 KeLP & University of Roma, Tor Vergata, Italy; Qatar Computing Research Institute,\\
 & HBKU, Qatar \cite{SemEval-2017:task3:KELP}\\
 MoRS & Universidade de Lisboa, Portugal\\
 & \cite{SemEval-2017:task3:LASIGUE}\\
 LearningToQuestion & Georgia Institute of Technology, Atlanta, GA, USA\\
 & \cite{SemEval-2017:task3:LEARNINGTOQUESTION}\\
 LS2N & LS2N\\
 & [no paper submitted] \\ %\cite{SemEval-2017:task3:LS2NSEMEVAL}
 NLM\_NIH & U.S. National Library of Medicine,
 Bethesda, MD, USA\\
 & \cite{SemEval-2017:task3:NLM-NIH}\\
 QU-BIGIR & Qatar University, Qatar\\
 & \cite{SemEval-2017:task3:QU-BIGIR}\\
 SCIR-QA & Harbin Institute of Technology, P.R. China\\
 & \cite{SemEval-2017:task3:SCIR-QA}\\
 SimBow & Orange Labs, France\\
 & \cite{SemEval-2017:task3:SIMBOW}\\
 SnowMan & Harbin Institute of Technology, P.R. China\\
 & [no paper submitted] \\ %\cite{SemEval-2017:task3:FA3L}
 SwissAlps & Zurich University of Applied Sciences, Switzerland\\
 & \cite{SemEval-2017:task3:SWISSALPS}\\
 TakeLab-QA & University of Zagreb, Croatia\\
 & \cite{SemEval-2017:task3:TAKELAB-QA}\\
 Talla  & Talla, Boston, MA, USA \\
 & \cite{SemEval-2017:task3:TALLA}\\
 TrentoTeam & University of Trento, Italy\\
 & \cite{SemEval-2017:task3:TRENTOTEAM}\\
 UINSUSKA-TiTech  & UIN Sultan Syarif Kasim Riau, Indonesia; Tokyo Institute of Technology, Japan \\
 & \cite{SemEval-2017:task3:UINSUSKA-TITECH}\\
 UPC-USMBA  & Universitat Polit\`{e}cnica de Catalunya, Spain; Sidi Mohamed Ben Abdellah University, Morocco \\
 & \cite{SemEval2017:task3:UPC-USMBA}\\
 \bottomrule
 \end{tabular}
 \caption{The participating teams and their affiliations.}
 \label{table:teams}
 \end{center}
 \end{table*}

%%%-------------------------------------------------
\subsection{Subtask A, English (Question-Comment Similarity)}

\begin{table*}[tbh]
\begin{center}
\begin{tabular}{clrrrrrrr}
\toprule
& \bf Submission & \bf MAP & \bf \scriptsize AvgRec & \bf \scriptsize MRR & \bf \scriptsize P & \bf \scriptsize R & \bf \scriptsize F1 & \bf \scriptsize Acc\\
\hline
\bf 1 & \bf KeLP-primary & \bf 88.43$_{1}$ & \bf \scriptsize 93.79$_{2}$ & \bf \scriptsize 92.82$_{1}$ & \bf \scriptsize 87.30$_{3}$ & \bf \scriptsize 58.24$_{9}$ & \bf \scriptsize 69.87$_{5}$ & \bf \scriptsize 73.89$_{3}$ \\
\bf 2 & \bf Beihang-MSRA-primary & \bf 88.24$_{2}$ & \bf \scriptsize 93.87$_{1}$ & \bf \scriptsize 92.34$_{2}$ & \bf \scriptsize 51.98$_{14}$ & \bf \scriptsize 100.00$_{1}$ & \bf \scriptsize 68.40$_{6}$ & \bf \scriptsize 51.98$_{13}$ \\
& Beihang-MSRA-contrastive2 & 88.18 & \scriptsize 93.91 & \scriptsize 92.45 & \scriptsize 51.98 & \scriptsize 100.00 & \scriptsize 68.40 & \scriptsize 51.98 \\
& Beihang-MSRA-contrastive1 & 88.17 & \scriptsize 93.82 & \scriptsize 92.17 & \scriptsize 51.98 & \scriptsize 100.00 & \scriptsize 68.40 & \scriptsize 51.98 \\
\bf 3 & \bf IIT-UHH-primary & \bf 86.88$_{3}$ & \bf \scriptsize 92.04$_{7}$ & \bf \scriptsize 91.20$_{5}$ & \bf \scriptsize 73.37$_{11}$ & \bf \scriptsize 74.52$_{3}$ & \bf \scriptsize 73.94$_{2}$ & \bf \scriptsize 72.70$_{4}$ \\
& ECNU-contrastive1 & 86.78 & \scriptsize 92.41 & \scriptsize 92.65 & \scriptsize 83.05 & \scriptsize 66.91 & \scriptsize 74.11 & \scriptsize 75.70 \\
\bf 4 & \bf ECNU-primary & \bf 86.72$_{4}$ & \bf \scriptsize 92.62$_{4}$ & \bf \scriptsize 91.45$_{3}$ & \bf \scriptsize 84.09$_{6}$ & \bf \scriptsize 72.16$_{4}$ & \bf \scriptsize 77.67$_{1}$ & \bf \scriptsize 78.43$_{1}$ \\
& EICA-contrastive2 & 86.60 & \scriptsize 92.25 & \scriptsize 90.67 & \scriptsize 88.50 & \scriptsize 31.32 & \scriptsize 46.27 & \scriptsize 62.18 \\
\bf 5 & \bf bunji-primary & \bf 86.58$_{5}$ & \bf \scriptsize 92.71$_{3}$ & \bf \scriptsize 91.37$_{4}$ & \bf \scriptsize 84.59$_{4}$ & \bf \scriptsize 63.43$_{5}$ & \bf \scriptsize 72.50$_{3}$ & \bf \scriptsize 74.98$_{2}$ \\
\bf 6 & \bf EICA-primary & \bf 86.53$_{6}$ & \bf \scriptsize 92.50$_{5}$ & \bf \scriptsize 89.57$_{8}$ & \bf \scriptsize 88.29$_{2}$ & \bf \scriptsize 30.20$_{12}$ & \bf \scriptsize 45.01$_{12}$ & \bf \scriptsize 61.64$_{11}$ \\
& EICA-contrastive1 & 86.48 & \scriptsize 92.18 & \scriptsize 90.69 & \scriptsize 88.43 & \scriptsize 29.61 & \scriptsize 44.37 & \scriptsize 61.40 \\
& IIT-UHH-contrastive1 & 86.35 & \scriptsize 91.74 & \scriptsize 91.40 & \scriptsize 79.42 & \scriptsize 51.94 & \scriptsize 62.80 & \scriptsize 68.02 \\
\bf 7 & \bf SwissAlps-primary & \bf 86.24$_{7}$ & \bf \scriptsize 92.28$_{6}$ & \bf \scriptsize 90.89$_{6}$ & \bf \scriptsize 90.78$_{1}$ & \bf \scriptsize 28.43$_{13}$ & \bf \scriptsize 43.30$_{13}$ & \bf \scriptsize 61.30$_{12}$ \\
& SwissAlps-contrastive1 & 85.53 & \scriptsize 91.98 & \scriptsize 90.52 & \scriptsize 90.37 & \scriptsize 24.03 & \scriptsize 37.97 & \scriptsize 59.18 \\
& bunji-contrastive1 & 85.29 & \scriptsize 91.77 & \scriptsize 91.48 & \scriptsize 83.14 & \scriptsize 56.34 & \scriptsize 67.16 & \scriptsize 71.37 \\
& IIT-UHH-contrastive2 & 85.24 & \scriptsize 91.37 & \scriptsize 90.38 & \scriptsize 81.22 & \scriptsize 57.65 & \scriptsize 67.43 & \scriptsize 71.06 \\
\bf 8 & \bf $^{\star}$FuRongWang-primary & \bf 84.26$_{8}$ & \bf \scriptsize 90.79$_{8}$ & \bf \scriptsize 89.40$_{9}$ & \bf \scriptsize 84.58$_{5}$ & \bf \scriptsize 48.98$_{10}$ & \bf \scriptsize 62.04$_{10}$ & \bf \scriptsize 68.84$_{7}$ \\
& bunji-contrastive2 & 84.01 & \scriptsize 90.45 & \scriptsize 89.17 & \scriptsize 81.88 & \scriptsize 59.03 & \scriptsize 68.60 & \scriptsize 71.91 \\
\bf 9 & \bf FA3L-primary & \bf 83.42$_{9}$ & \bf \scriptsize 89.90$_{9}$ & \bf \scriptsize 90.32$_{7}$ & \bf \scriptsize 73.82$_{10}$ & \bf \scriptsize 59.62$_{6}$ & \bf \scriptsize 65.96$_{9}$ & \bf \scriptsize 68.02$_{8}$ \\
& ECNU-contrastive2 & 83.15 & \scriptsize 90.01 & \scriptsize 89.46 & \scriptsize 75.06 & \scriptsize 78.86 & \scriptsize 76.91 & \scriptsize 75.39 \\
& LS2N-contrastive2 & 82.91 & \scriptsize 89.70 & \scriptsize 89.58 & \scriptsize 72.19 & \scriptsize 71.77 & \scriptsize 71.98 & \scriptsize 70.96 \\
& FA3L-contrastive1 & 82.87 & \scriptsize 89.64 & \scriptsize 89.98 & \scriptsize 77.28 & \scriptsize 56.27 & \scriptsize 65.12 & \scriptsize 68.67 \\
& SnowMan-contrastive1 & 82.01 & \scriptsize 89.36 & \scriptsize 88.56 & \scriptsize 75.92 & \scriptsize 73.47 & \scriptsize 74.67 & \scriptsize 74.10 \\
\bf 10 & \bf SnowMan-primary & \bf 81.84$_{10}$ & \bf \scriptsize 88.67$_{10}$ & \bf \scriptsize 87.21$_{12}$ & \bf \scriptsize 79.54$_{8}$ & \bf \scriptsize 58.44$_{7}$ & \bf \scriptsize 67.37$_{7}$ & \bf \scriptsize 70.58$_{5}$ \\
\bf 11 & \bf TakeLab-QA-primary & \bf 81.14$_{11}$ & \bf \scriptsize 88.48$_{12}$ & \bf \scriptsize 87.51$_{11}$ & \bf \scriptsize 78.72$_{9}$ & \bf \scriptsize 58.31$_{8}$ & \bf \scriptsize 66.99$_{8}$ & \bf \scriptsize 70.14$_{6}$ \\
\bf 12 & \bf LS2N-primary & \bf 80.99$_{12}$ & \bf \scriptsize 88.55$_{11}$ & \bf \scriptsize 87.92$_{10}$ & \bf \scriptsize 80.07$_{7}$ & \bf \scriptsize 43.27$_{11}$ & \bf \scriptsize 56.18$_{11}$ & \bf \scriptsize 64.91$_{10}$ \\
& TakeLab-QA-contrastive1 & 79.71 & \scriptsize 87.31 & \scriptsize 87.03 & \scriptsize 73.88 & \scriptsize 62.77 & \scriptsize 67.87 & \scriptsize 69.11 \\
& TakeLab-QA-contrastive2 & 78.98 & \scriptsize 86.33 & \scriptsize 87.13 & \scriptsize 80.06 & \scriptsize 56.66 & \scriptsize 66.36 & \scriptsize 70.14 \\
\bf 13 & \bf TrentoTeam-primary & \bf 78.56$_{13}$ & \bf \scriptsize 86.66$_{13}$ & \bf \scriptsize 85.76$_{13}$ & \bf \scriptsize 65.59$_{12}$ & \bf \scriptsize 75.71$_{2}$ & \bf \scriptsize 70.28$_{4}$ & \bf \scriptsize 66.72$_{9}$ \\
& LS2N-contrastive1 & 74.08 & \scriptsize 81.88 & \scriptsize 81.66 & \scriptsize 70.66 & \scriptsize 28.30 & \scriptsize 40.41 & \scriptsize 56.62 \\
\bf 14 & \bf MoRS-primary & \bf 63.32$_{14}$ & \bf \scriptsize 71.67$_{14}$ & \bf \scriptsize 71.99$_{14}$ & \bf \scriptsize 59.23$_{13}$ & \bf \scriptsize  5.06$_{14}$ & \bf \scriptsize  9.32$_{14}$ & \bf \scriptsize 48.84$_{14}$ \\
\midrule
& Baseline 1 (chronological) & \bf 72.61 & \scriptsize \bf 79.32 & \scriptsize \bf 82.37 & \scriptsize  --- & \scriptsize  --- & \scriptsize  --- & \scriptsize --- \\
& Baseline 2 (random) & 62.30 & \scriptsize 70.56 & \scriptsize 68.74 & \scriptsize 53.15 & \scriptsize 75.97 & \scriptsize 62.54 & \scriptsize \bf 52.70 \\
& Baseline 3 (all `true') & --- & \scriptsize --- & \scriptsize --- & \scriptsize 51.98 & \scriptsize 100.00 & \scriptsize \bf 68.40 & \scriptsize 51.98 \\
& Baseline 4 (all `false') & --- & \scriptsize --- & \scriptsize --- & \scriptsize  --- & \scriptsize  --- & \scriptsize  --- & \scriptsize 48.02 \\
\bottomrule
\end{tabular}
\caption{\textbf{Subtask A, English (Question-Comment Similarity)}: results for all submissions. The first column shows the rank of the primary runs with respect to the official MAP score. The second column contains the team's name and its submission type (primary vs. contrastive).
The following columns show the results for the primary, and then for other, unofficial evaluation measures. The subindices show the rank of the primary runs with respect to the evaluation measure in the respective column. All results are presented as percentages.
The system marked with a $^{\star}$ was a late submission.}
\label{table:results:subtaskA}
\end{center}
\end{table*}

Table~\ref{table:results:subtaskA} shows the results for subtask A, English, which attracted 14 teams (two more than in the 2016 edition). In total 31 runs were submitted: 14 primary and 17 contrastive. 
The last four rows of the table show the performance of four baselines.
The first one is the chronological ranking, where the comments are ordered by their time of posting; we can see that all submissions but one outperform this baseline on all three ranking measures.
The second baseline is a random baseline, which is 10 MAP points below the chronological ranking. 
Baseline 3 classifies all comments as Good, and it outperforms all but three of the primary systems in terms of F$_1$ and one system in terms of Accuracy. However, it should be noted that the systems were not optimized for such measures.
Finally, baseline 4 classifies all comments as Bad; it is outperformed by all primary systems in terms of Accuracy. 

The winner of Subtask A is \emph{KeLP} with a MAP of 88.43, closely followed by \emph{Beihang-MSRA}, scoring 88.24. Relatively far from the first two, we find five systems, \emph{IIT-UHH},
\emph{ECNU}, \emph{bunji}, \emph{EICA} and \emph{SwissAlps}, which all obtained an MAP of around 86.5.

\subsection{Subtask B, English (Question-Question Similarity)}
Table~\ref{table:results:subtaskB} shows the results for subtask B, English, which attracted 13 teams (3 more than in last year's edition) and 34 runs:~13 primary and 21 contrastive. This is known to be a hard task. In contrast to the 2016 results, 
%KMV: how many systems were beat out by the baseline last year? You may have said it but I don't remember. Probably worth stating here, as follows:
 in which only 6 out of 11 teams beat %KMV add in "X" ---> OK FIXED
the strong IR baseline (i.e., ordering the related questions in the order provided by the search engine), this year 10 % outperforms only 3 
of the 13 systems outperformed this baseline 
in terms of MAP, AvgRec and MRR. Moreover, the improvements for the best systems over the IR baseline are larger (reaching $>7$ MAP points absolute). This is a remarkable improvement over last year's results.

\noindent The random baseline outperforms two systems in terms of Accuracy.
The ``all-good'' baseline is below almost all systems on F$_1$, but the ``all-false'' baseline yields the best Accuracy results. This is partly because the label distribution in the dataset is biased (81.5\% of negative cases), but also because the systems were optimized for MAP rather than for classification accuracy (or precision/recall). 
%Actually, several systems did not even provide %sensible
%classification predictions, as they were optional in all subtasks.
%KMV: I'm wondering if you need to say more about this. How do you measure P/R/F if you don't have these classification predictions? --> REMOVED the sentence above, it is probably related to Task D and it is not really meaningful here.

The winner of the task is \emph{SimBow} with a MAP of 47.22, followed by \emph{LearningToQuestion} with 46.93, \emph{KeLP} with 46.66, and \emph{Talla} with 45.70. The other nine systems scored sensibly lower than them, ranging from about 41 to 45. 
Note that the contrastive1 run of \emph{KeLP}, which corresponds to the \emph{KeLP} system from last year \cite{SemEval2016:task3:KeLP}, achieved an even higher MAP of 49.00. 
%This year, they used a text selection techniques proposed in \cite{RomeoMBMBHZMG16,DBLP:conf/coling/Barron-CedenoMR16} for their primary submission. % in the 2016 system.  %KMV: I'm confused, I assume that the "new" techniques are new in 2017? I just commented this out but make sure that's right. --> the technique was  designed after SemEval and thus is new for SemEval 2017.
%In their SemEval system paper, \newcite{SemEval-2017:task3:KELP} explain that this approach is sensitive to data distribution; the differences between this and last year's data require some improvement of their proposed approach.

\begin{table*}[tbh]
\begin{center}
\begin{tabular}{clrrrrrrr}
\toprule
& \bf Submission & \bf MAP  & \bf \scriptsize AvgRec & \bf \scriptsize MRR & \bf \scriptsize P & \bf \scriptsize R & \bf \scriptsize F1 & \bf \scriptsize Acc\\
\midrule
& KeLP-contrastive1 & 49.00 & \scriptsize 83.92 & \scriptsize 52.41 & \scriptsize 36.18 & \scriptsize 88.34 & \scriptsize 51.34 & \scriptsize 68.98 \\
& SimBow-contrastive2 & 47.87 & \scriptsize 82.77 & \scriptsize 50.97 & \scriptsize 27.03 & \scriptsize 93.87 & \scriptsize 41.98 & \scriptsize 51.93 \\
\bf 1 & \bf SimBow-primary & \bf 47.22$_{1}$ & \bf \scriptsize 82.60$_{1}$ & \bf \scriptsize 50.07$_{3}$ & \bf \scriptsize 27.30$_{10}$ & \bf \scriptsize 94.48$_{3}$ & \bf \scriptsize 42.37$_{9}$ & \bf \scriptsize 52.39$_{11}$ \\
& LearningToQuestion-contrastive2 & 47.20 & \scriptsize 81.73 & \scriptsize 53.22 & \scriptsize 18.52 & \scriptsize 100.00 & \scriptsize 31.26 & \scriptsize 18.52 \\
& LearningToQuestion-contrastive1 & 47.03 & \scriptsize 81.45 & \scriptsize 52.47 & \scriptsize 18.52 & \scriptsize 100.00 & \scriptsize 31.26 & \scriptsize 18.52 \\
\bf 2 & \bf LearningToQuestion-primary & \bf 46.93$_{2}$ & \bf \scriptsize 81.29$_{4}$ & \bf \scriptsize 53.01$_{1}$ & \bf \scriptsize 18.52$_{12}$ & \bf \scriptsize 100.00$_{1}$ & \bf \scriptsize 31.26$_{12}$ & \bf \scriptsize 18.52$_{12}$ \\
& SimBow-contrastive1 & 46.84 & \scriptsize 82.73 & \scriptsize 50.43 & \scriptsize 27.80 & \scriptsize 94.48 & \scriptsize 42.96 & \scriptsize 53.52 \\
\bf 3 & \bf KeLP-primary & \bf 46.66$_{3}$ & \bf \scriptsize 81.36$_{3}$ & \bf \scriptsize 50.85$_{2}$ & \bf \scriptsize 36.01$_{3}$ & \bf \scriptsize 85.28$_{5}$ & \bf \scriptsize 50.64$_{1}$ & \bf \scriptsize 69.20$_{5}$ \\
& Talla-contrastive1 & 46.54 & \scriptsize 82.15 & \scriptsize 49.61 & \scriptsize 30.39 & \scriptsize 76.07 & \scriptsize 43.43 & \scriptsize 63.30 \\
& Talla-contrastive2 & 46.31 & \scriptsize 81.81 & \scriptsize 49.14 & \scriptsize 29.88 & \scriptsize 74.23 & \scriptsize 42.61 & \scriptsize 62.95 \\
\bf 4 & \bf Talla-primary & \bf 45.70$_{4}$ & \bf \scriptsize 81.48$_{2}$ & \bf \scriptsize 49.55$_{5}$ & \bf \scriptsize 29.59$_{9}$ & \bf \scriptsize 76.07$_{8}$ & \bf \scriptsize 42.61$_{8}$ & \bf \scriptsize 62.05$_{8}$ \\
& Beihang-MSRA-contrastive2 & 44.79 & \scriptsize 79.13 & \scriptsize 49.89 & \scriptsize 18.52 & \scriptsize 100.00 & \scriptsize 31.26 & \scriptsize 18.52 \\
\bf 5 & \bf Beihang-MSRA-primary & \bf 44.78$_{5}$ & \bf \scriptsize 79.13$_{7}$ & \bf \scriptsize 49.88$_{4}$ & \bf \scriptsize 18.52$_{13}$ & \bf \scriptsize 100.00$_{2}$ & \bf \scriptsize 31.26$_{13}$ & \bf \scriptsize 18.52$_{13}$ \\
& NLM\_NIH-contrastive1 & 44.66 & \scriptsize 79.66 & \scriptsize 48.08 & \scriptsize 33.68 & \scriptsize 79.14 & \scriptsize 47.25 & \scriptsize 67.27 \\
\bf 6 & \bf NLM\_NIH-primary & \bf 44.62$_{6}$ & \bf \scriptsize 79.59$_{5}$ & \bf \scriptsize 47.74$_{6}$ & \bf \scriptsize 33.68$_{5}$ & \bf \scriptsize 79.14$_{6}$ & \bf \scriptsize 47.25$_{3}$ & \bf \scriptsize 67.27$_{6}$ \\
& UINSUSKA-TiTech-contrastive1 & 44.29 & \scriptsize 78.59 & \scriptsize 48.97 & \scriptsize 34.47 & \scriptsize 68.10 & \scriptsize 45.77 & \scriptsize 70.11 \\
& NLM\_NIH-contrastive2 & 44.29 & \scriptsize 79.05 & \scriptsize 47.45 & \scriptsize 33.68 & \scriptsize 79.14 & \scriptsize 47.25 & \scriptsize 67.27 \\
& Beihang-MSRA-contrastive1 & 43.89 & \scriptsize 79.48 & \scriptsize 48.18 & \scriptsize 18.52 & \scriptsize 100.00 & \scriptsize 31.26 & \scriptsize 18.52 \\
\bf 7 & \bf UINSUSKA-TiTech-primary & \bf 43.44$_{7}$ & \bf \scriptsize 77.50$_{11}$ & \bf \scriptsize 47.03$_{9}$ & \bf \scriptsize 35.71$_{4}$ & \bf \scriptsize 67.48$_{11}$ & \bf \scriptsize 46.71$_{4}$ & \bf \scriptsize 71.48$_{4}$ \\
\bf 8 & \bf IIT-UHH-primary & \bf 43.12$_{8}$ & \bf \scriptsize 79.23$_{6}$ & \bf \scriptsize 47.25$_{7}$ & \bf \scriptsize 26.85$_{11}$ & \bf \scriptsize 71.17$_{10}$ & \bf \scriptsize 38.99$_{10}$ & \bf \scriptsize 58.75$_{10}$ \\
& UINSUSKA-TiTech-contrastive2 & 43.06 & \scriptsize 76.45 & \scriptsize 46.22 & \scriptsize 35.71 & \scriptsize 67.48 & \scriptsize 46.71 & \scriptsize 71.48 \\
\bf 9 & \bf SCIR-QA-primary & \bf 42.72$_{9}$ & \bf \scriptsize 78.24$_{9}$ & \bf \scriptsize 46.65$_{10}$ & \bf \scriptsize 31.26$_{8}$ & \bf \scriptsize 89.57$_{4}$ & \bf \scriptsize 46.35$_{5}$ & \bf \scriptsize 61.59$_{9}$ \\
& SCIR-QA-contrastive1 & 42.72 & \scriptsize 78.24 & \scriptsize 46.65 & \scriptsize 32.69 & \scriptsize 83.44 & \scriptsize 46.98 & \scriptsize 65.11 \\
& ECNU-contrastive2 & 42.48 & \scriptsize 79.44 & \scriptsize 45.09 & \scriptsize 36.47 & \scriptsize 78.53 & \scriptsize 49.81 & \scriptsize 70.68 \\
& IIT-UHH-contrastive2 & 42.38 & \scriptsize 78.59 & \scriptsize 46.82 & \scriptsize 32.99 & \scriptsize 59.51 & \scriptsize 42.45 & \scriptsize 70.11 \\
& ECNU-contrastive1 & 42.37 & \scriptsize 78.41 & \scriptsize 45.04 & \scriptsize 34.34 & \scriptsize 83.44 & \scriptsize 48.66 & \scriptsize 67.39 \\
& IIT-UHH-contrastive1 & 42.29 & \scriptsize 78.41 & \scriptsize 46.40 & \scriptsize 32.66 & \scriptsize 59.51 & \scriptsize 42.17 & \scriptsize 69.77 \\
\bf 10 & \bf FA3L-primary & \bf 42.24$_{10}$ & \bf \scriptsize 77.71$_{10}$ & \bf \scriptsize 47.05$_{8}$ & \bf \scriptsize 33.17$_{6}$ & \bf \scriptsize 40.49$_{13}$ & \bf \scriptsize 36.46$_{11}$ & \bf \scriptsize 73.86$_{2}$ \\
& LS2N-contrastive1 & 42.06 & \scriptsize 77.36 & \scriptsize 47.13 & \scriptsize 32.01 & \scriptsize 59.51 & \scriptsize 41.63 & \scriptsize 69.09 \\
\bf 11 & \bf ECNU-primary & \bf 41.37$_{11}$ & \bf \scriptsize 78.71$_{8}$ & \bf \scriptsize 44.52$_{13}$ & \bf \scriptsize 37.43$_{1}$ & \bf \scriptsize 76.69$_{7}$ & \bf \scriptsize 50.30$_{2}$ & \bf \scriptsize 71.93$_{3}$ \\
\bf 12 & \bf EICA-primary & \bf 41.11$_{12}$ & \bf \scriptsize 77.45$_{12}$ & \bf \scriptsize 45.57$_{12}$ & \bf \scriptsize 32.60$_{7}$ & \bf \scriptsize 72.39$_{9}$ & \bf \scriptsize 44.95$_{6}$ & \bf \scriptsize 67.16$_{7}$ \\
& EICA-contrastive1 & 41.07 & \scriptsize 77.70 & \scriptsize 46.38 & \scriptsize 32.30 & \scriptsize 70.55 & \scriptsize 44.32 & \scriptsize 67.16 \\
\bf 13 & \bf LS2N-primary & \bf 40.56$_{13}$ & \bf \scriptsize 76.67$_{13}$ & \bf \scriptsize 46.33$_{11}$ & \bf \scriptsize 36.55$_{2}$ & \bf \scriptsize 53.37$_{12}$ & \bf \scriptsize 43.39$_{7}$ & \bf \scriptsize 74.20$_{1}$ \\
& EICA-contrastive2 & 40.04 & \scriptsize 76.98 & \scriptsize 44.00 & \scriptsize 31.69 & \scriptsize 71.17 & \scriptsize 43.86 & \scriptsize 66.25 \\
\midrule
& Baseline 1 (IR) & \bf 41.85 & \scriptsize \bf 77.59 & \scriptsize \bf 46.42 & \scriptsize  --- & \scriptsize  --- & \scriptsize  --- & \scriptsize --- \\
& Baseline 2 (random) & 29.81 & \scriptsize 62.65 & \scriptsize 33.02 & \scriptsize 18.72 & \scriptsize 75.46 & \scriptsize 30.00 & \scriptsize 34.77 \\
& Baseline 3 (all `true') & --- & \scriptsize --- & \scriptsize --- & \scriptsize 18.52 & \scriptsize 100.00 & \scriptsize \bf 31.26 & \scriptsize 18.52 \\
& Baseline 4 (all `false') & --- & \scriptsize --- & \scriptsize --- & \scriptsize  --- & \scriptsize  --- & \scriptsize  --- & \scriptsize \bf 81.48 \\
\bottomrule
\end{tabular}
\caption{\textbf{Subtask B, English (Question-Question Similarity):} results for all submissions. The first column shows the rank of the primary runs with respect to the official MAP score. The second column contains the team's name and its submission type (primary vs. contrastive).
The following columns show the results for the primary, and then for other, unofficial evaluation measures. The subindices show the rank of the primary runs with respect to the evaluation measure in the respective column. All results are presented as percentages.}
\label{table:results:subtaskB}
\end{center}
\end{table*}

%%%-------------------------------------------------
\subsection{Subtask C, English (Question-External Comment Similarity)}

The results for subtask C, English are shown in Table~\ref{table:results:subtaskC}. This subtask attracted 6 teams (sizable decrease compared to last year's 10 teams), and 14 runs: 6 primary and 8 contrastive. 
The test set from 2017 had much more skewed label distribution, with only 2.8\% positive instances, compared to the $\sim$10\% of the 2016 test set. This makes the overall MAP scores look much lower, as the number of examples without a single positive comment increased significantly, and they contribute 0 to the average, due to the definition of the measure. Consequently, the results cannot be compared directly to last year's. 
%In relative terms, a similar pattern is observed. 

All primary systems managed to outperform all baselines with respect to the ranking measures. Moreover, all but one system outperformed the ``all true'' system on F$_1$, and all of them were below the accuracy of the ``all false'' baseline, due to the extreme class imbalance. 

The best-performing team for subtask C is \emph{IIT-UHH}, with a MAP of 15.46, followed by \emph{bunji} with 14.71, and \emph{KeLP} with 14.35. The contrastive1 run of \emph{bunji}, which used a neural network, obtained the highest MAP, 16.57, two points higher than their primary run, which also uses the comment plausibility features. Thus, the difference seems to be due to the use of comment plausibility features, which hurt the accuracy. In their SemEval system paper, \newcite{SemEval-2017:task3:BUNJI} explain  
%KMV: cite?
that the similarity features are more important for Subtask C than plausibility features.

\noindent Indeed, Subtask C contains many comments that are not related to the original question, while candidate comments for subtask A are almost always on the same topic. Another explanation may be the overfitting to the development set since the authors manually designed plausibility features using that set. As a result, such features perform much worse on the 2017 test set.

\begin{table*}[tbh]
\begin{center}
\begin{tabular}{clrrrrrrr}
\toprule
& \bf Submission & \bf MAP & \bf \scriptsize AvgRec & \bf \scriptsize MRR & \bf \scriptsize P & \bf \scriptsize R & \bf \scriptsize F1 & \bf \scriptsize Acc\\
\hline
& bunji-contrastive2 & 16.57 & \scriptsize 30.98 & \scriptsize 17.04 & \scriptsize 19.83 & \scriptsize 19.11 & \scriptsize 19.46 & \scriptsize 95.58 \\
\bf 1 & \bf IIT-UHH-primary & \bf 15.46$_{1}$ & \bf \scriptsize 33.42$_{1}$ & \bf \scriptsize 18.14$_{1}$ & \bf \scriptsize  8.41$_{3}$ & \bf \scriptsize 51.22$_{3}$ & \bf \scriptsize 14.44$_{2}$ & \bf \scriptsize 83.03$_{4}$ \\
& IIT-UHH-contrastive1 & 15.43 & \scriptsize 33.78 & \scriptsize 17.52 & \scriptsize  9.45 & \scriptsize 54.07 & \scriptsize 16.08 & \scriptsize 84.23 \\
\bf 2 & \bf bunji-primary & \bf 14.71$_{2}$ & \bf \scriptsize 29.47$_{4}$ & \bf \scriptsize 16.48$_{2}$ & \bf \scriptsize 20.26$_{1}$ & \bf \scriptsize 19.11$_{4}$ & \bf \scriptsize 19.67$_{1}$ & \bf \scriptsize 95.64$_{2}$ \\
& EICA-contrastive1 & 14.60 & \scriptsize 32.71 & \scriptsize 16.14 & \scriptsize 10.80 & \scriptsize  9.35 & \scriptsize 10.02 & \scriptsize 95.31 \\
\bf 3 & \bf KeLP-primary & \bf 14.35$_{3}$ & \bf \scriptsize 30.74$_{2}$ & \bf \scriptsize 16.07$_{3}$ & \bf \scriptsize  6.48$_{5}$ & \bf \scriptsize 89.02$_{2}$ & \bf \scriptsize 12.07$_{4}$ & \bf \scriptsize 63.75$_{5}$ \\
& IIT-UHH-contrastive2 & 14.00 & \scriptsize 30.53 & \scriptsize 14.65 & \scriptsize  5.98 & \scriptsize 85.37 & \scriptsize 11.17 & \scriptsize 62.06 \\
\bf 4 & \bf EICA-primary & \bf 13.48$_{4}$ & \bf \scriptsize 24.44$_{6}$ & \bf \scriptsize 16.04$_{4}$ & \bf \scriptsize  7.69$_{4}$ & \bf \scriptsize  0.41$_{6}$ & \bf \scriptsize  0.77$_{6}$ & \bf \scriptsize 97.08$_{1}$ \\
& ECNU-contrastive2 & 13.29 & \scriptsize 30.15 & \scriptsize 14.95 & \scriptsize 13.86 & \scriptsize 26.42 & \scriptsize 18.18 & \scriptsize 93.35 \\
\bf 5 & \bf $^{\star}$FuRongWang-primary & \bf 13.23$_{5}$ & \bf \scriptsize 29.51$_{3}$ & \bf \scriptsize 14.27$_{5}$ & \bf \scriptsize  2.80$_{6}$ & \bf \scriptsize 100.00$_{1}$ & \bf \scriptsize  5.44$_{5}$ & \bf \scriptsize  2.80$_{6}$ \\
& EICA-contrastive2 & 13.18 & \scriptsize 25.16 & \scriptsize 15.05 & \scriptsize 10.00 & \scriptsize  0.81 & \scriptsize  1.50 & \scriptsize 97.02 \\
\bf 6 & \bf ECNU-primary & \bf 10.54$_{6}$ & \bf \scriptsize 25.56$_{5}$ & \bf \scriptsize 11.09$_{6}$ & \bf \scriptsize 13.44$_{2}$ & \bf \scriptsize 13.82$_{5}$ & \bf \scriptsize 13.63$_{3}$ & \bf \scriptsize 95.10$_{3}$ \\
& ECNU-contrastive1 & 10.54 & \scriptsize 25.56 & \scriptsize 11.09 & \scriptsize 13.83 & \scriptsize 14.23 & \scriptsize 14.03 & \scriptsize 95.13 \\
& bunji-contrastive1 &  8.19 & \scriptsize 15.12 & \scriptsize  9.25 & \scriptsize  0.00 & \scriptsize  0.00 & \scriptsize  0.00 & \scriptsize 97.20 \\
\midrule
& Baseline 1 (IR) &  \bf 9.18 & \scriptsize \bf 21.72 & \scriptsize \bf 10.11 & \scriptsize  --- & \scriptsize --- & \scriptsize  --- & \scriptsize --- \\
& Baseline 2 (random) &  5.77 & \scriptsize 7.69 & \scriptsize 5.70 & \scriptsize  2.76 & \scriptsize 73.98 & \scriptsize  5.32 & \scriptsize 26.37 \\
& Baseline 3 (all `true') &  --- & \scriptsize  --- & \scriptsize  --- & \scriptsize  2.80 & \scriptsize 100.00 & \scriptsize \bf 5.44 & \scriptsize  2.80 \\
& Baseline 4 (all `false') &  --- & \scriptsize  --- & \scriptsize  --- & \scriptsize  --- & \scriptsize  --- & \scriptsize  --- & \scriptsize \bf 97.20 \\
\bottomrule
\end{tabular}
\caption{\textbf{Subtask C, English (Question-External Comment Similarity):} results for all submissions. The first column shows the rank of the primary runs with respect to the official MAP score. The second column contains the team's name and its submission type (primary vs. contrastive). The following columns show the results for the primary, and then for other, unofficial evaluation measures. The subindices show the rank of the primary runs with respect to the evaluation measure in the respective column. All results are presented as percentages.
The system marked with a $^{\star}$ was a late submission.}
\label{table:results:subtaskC}
\end{center}
\end{table*}

%%%-------------------------------------------------
\subsection{Subtask D, Arabic (Reranking the Correct Answers for a New Question)}

Finally, the results for subtask D, Arabic are shown in Table~\ref{table:results:subtaskD}. This year, subtask D attracted only 3 teams, which submitted 6 runs: 3 primary and 3 contrastive. 
Compared to last year, the 2017 test set contains a significantly larger number of positive question--answer pairs ($\sim$40\% in 2017, compared to $\sim$20\% in 2016), and thus the MAP scores are higher this year. Moreover, this year, the IR baseline is coming from Google and is thus very strong and difficult to beat. Indeed, only the best system was able to improve on it (marginally) in terms of MAP, MRR and AvgRec. 

As in some of the other tasks, the participants in Subtask D did not concentrate on optimizing for precision/recall/F$_1$/accuracy and they did not produce sensible class predictions in most cases. 
%Thus, it is not worthwhile to discuss the last four columns in Table~\ref{table:results:subtaskD}. 

The best-performing system is \emph{GW\_QA} with a MAP score of 61.16, which barely improves over the IR baseline of 60.55. The other two systems \emph{UPC-USMBA} and \emph{QU\_BIGIR} are about 3-4 points behind.

\begin{table*}[tbh]
\begin{center}
\begin{tabular}{clrrrrrrr}
\toprule
& \bf Submission & \bf MAP  & \bf \scriptsize AvgRec & \bf \scriptsize MRR & \bf \scriptsize P & \bf \scriptsize R & \bf \scriptsize F1 & \bf \scriptsize Acc\\
\midrule
\bf 1 & \bf GW\_QA-primary & \bf 61.16$_{1}$ & \bf \scriptsize 85.43$_{1}$ & \bf \scriptsize 66.85$_{1}$ & \bf \scriptsize  0.00$_{3}$ & \bf \scriptsize  0.00$_{3}$ & \bf \scriptsize  0.00$_{3}$ & \bf \scriptsize 60.77$_{2}$ \\
& QU\_BIGIR-contrastive2 & 59.48 & \scriptsize 83.83 & \scriptsize 64.56 & \scriptsize 55.35 & \scriptsize 70.95 & \scriptsize 62.19 & \scriptsize 66.15 \\
& QU\_BIGIR-contrastive1 & 59.13 & \scriptsize 83.56 & \scriptsize 64.68 & \scriptsize 49.37 & \scriptsize 85.41 & \scriptsize 62.57 & \scriptsize 59.91 \\
\bf 2 & \bf UPC-USMBA-primary & \bf 57.73$_{2}$ & \bf \scriptsize 81.76$_{3}$ & \bf \scriptsize 62.88$_{2}$ & \bf \scriptsize 63.41$_{1}$ & \bf \scriptsize 33.00$_{2}$ & \bf \scriptsize 43.41$_{2}$ & \bf \scriptsize 66.24$_{1}$ \\
\bf 3 & \bf QU\_BIGIR-primary & \bf 56.69$_{3}$ & \bf \scriptsize 81.89$_{2}$ & \bf \scriptsize 61.83$_{3}$ & \bf \scriptsize 41.59$_{2}$ & \bf \scriptsize 70.16$_{1}$ & \bf \scriptsize 52.22$_{1}$ & \bf \scriptsize 49.64$_{3}$ \\
& UPC-USMBA-contrastive1 & 56.66 & \scriptsize 81.16 & \scriptsize 62.87 & \scriptsize 45.00 & \scriptsize 64.04 & \scriptsize 52.86 & \scriptsize 55.18 \\
\midrule
& Baseline 1 (IR) & \bf 60.55 & \scriptsize \bf 85.06 & \scriptsize \bf 66.80 & \scriptsize --- & \scriptsize --- & \scriptsize --- & \scriptsize --- \\
& Baseline 2 (random) & 48.48 & \scriptsize 73.89 & \scriptsize 53.27 & \scriptsize 39.04 & \scriptsize 66.43 & \scriptsize 49.18 & \scriptsize 46.13 \\
& Baseline 3 (all `true') & --- & \scriptsize --- & \scriptsize --- & \scriptsize 39.23 & \scriptsize 100.00 & \scriptsize \bf 56.36 & \scriptsize 39.23 \\
& Baseline 4 (all `false') & --- & \scriptsize --- & \scriptsize --- & \scriptsize  --- & \scriptsize  --- & \scriptsize  --- & \scriptsize \bf 60.77 \\
\bottomrule
\end{tabular}
\caption{\textbf{Subtask D, Arabic (Reranking the correct answers for a new question):} results for all submissions. The first column shows the rank of the primary runs with respect to the official MAP score. The second column contains the team's name and its submission type (primary vs. contrastive). The following columns show the results for the primary, and then for other, unofficial evaluation measures. The subindices show the rank of the primary runs with respect to the evaluation measure in the respective column. All results are presented as percentages.}
\label{table:results:subtaskD}
\end{center}
\end{table*}

%%%-------------------------------------------------
\subsection{Subtask E, English (Multi-Domain Question Duplicate Detection)}

The baselines for Subtask E can be found in \tabref{tab:baseline-e}. The IR baseline is BM25 with perfect truncation after the final relevant document for a given document (equating to an empty result list if there are no relevant documents). The zero results baseline is the score for a system that returns an empty result list for every single query. This is a high number for each subforum because for many queries there are no duplicate questions in the archive.

As previously stated, there are no results submitted by participants to be discussed for this subtask. Eight teams signed up to participate, but unfortunately none of them submitted test results.

\begin{table*}[t]
\centering
\renewcommand{\arraystretch}{1.1}
%\setlength\tabcolsep{5pt} % make LaTeX figure out width of inter-column spaces
%\begin{tabular}{l *{1}{S[table-format=1.4]}}
\begin{tabular}{l c}
\toprule
\bf Baseline & \bf TMAP \\
\midrule
Android Baseline 1 (IR oracle) & 99.00 \\
Android Baseline 2 (all empty results) & 98.56 \\[0.5ex]
English Baseline 1 (IR oracle) & 98.05 \\
English Baseline 2 (all empty results) & 97.65 \\[0.5ex]
Gaming Baseline 1 (IR oracle) & 99.18 \\
Gaming Baseline 2 (all empty results) & 98.73 \\[0.5ex]
Wordpress Baseline 1 (IR oracle) & 99.21 \\
Wordpress Baseline 2 (all empty results) & 98.98 \\
\bottomrule
\end{tabular}
\caption{\textbf{Subtask E, English (Multi-Domain Duplicate Detection):} Baseline results on the test dataset. The empty result baseline has an empty result list for all queries. The IR baselines are the results of applying BM25 with perfect truncation. All results are presented as percentages.}
\label{tab:baseline-e}
\end{table*}

%\input{results_subtask_E.tex}

%\input{participants-results.tex}

%%%----------------- D i s c u s s i o n
\section{Discussion and Conclusions}
\label{sec:discussion}
%\alex{Alessandro. Done!}

In this section, we first describe features that are common across the different subtasks. Then, we discuss the characteristics of the best systems for each subtask with focus on the machine learning algorithms and the instance representations used.

\subsection{Feature Types}

The features the participants used across the sutbtasks can be organized into the following groups:

(\emph{i})~\emph{similarity features} between questions and comments from their threads or between original questions and related questions, e.g., cosine similarity applied to lexical, syntactic and semantic representations, including distributed representations, often derived using neural networks;

(\emph{ii})~\emph{content features}, which are special signals that can clearly indicate a bad comment, e.g., when a comment contains ``thanks''; %KMV: hmmm, why does "thanks" indicate a bad answer? --> because it means that the answer is from someone who asked something instead of providing something. I changed answer with comment so it becomes clearer.

(\emph{iii})~\emph{thread level/meta features}, e.g., user ID, comment rank in the thread;

(\emph{iv})~\emph{automatically generated features} from syntactic structures using tree kernels.

Generally, similarity features were developed for the subtasks as follows:

\paragraph{Subtask A.} Similarities between question subject vs.~comment, question body vs.~comment, and question subject+body vs.~comment.

\paragraph{Subtask B.} Similarities between the original and the related question at different levels: subject vs.~subject, body vs.~body, and subject+body vs.~subject+body.

\paragraph{Subtask C.} The same as above, plus the similarities of the original question, subject and body at all levels with the comments from the thread of the related question.

\paragraph{Subtask D.} The same as above, without information about the thread, as there is no thread.\\

The similarity scores to be used as features were computed in various ways, e.g., most teams used dot product calculated over word $n$-grams ($n$=1,2,3), character $n$-grams, or with TF-IDF weighting. Simple word overlap, i.e., the number of common words between two texts, was also considered, often normalized, e.g., by question/comment length. Overlap in terms of nouns or named entities was also explored.

\subsection{Learning Methods}
% In contrast with the previous two editions, this time many more creative learning methods were used in addition to SVMs, which still yield the best performance overall. Indeed, KeLP, the system that won Subtask A (and achieved the best performance on Subtask B) this year, was methodologically %KMV: I added "methodologically", is that right?, --> yes it is more precise as it was retrained.
% the same as in 2016. Its main feature was again
% tree kernels with relational links in SVMs  \cite{SemEval2016:task3:KeLP}. More specifically, questions are aligned with comments (or with other questions) by means of a special REL tag, directly annotated in the parse trees. They used the homonymous tree kernel software toolkit, KeLP \cite{filice-EtAl:2015:KeLP2015} to learn such relational structural models. They further used some standard text similarity measures.
% %
% %KMV: I'm a bit confused here. Are all these cites about KeLP? --> yes, some info more is needed, I rewrote the paragraph above.
% Variants of this approach were successfully used in related research, e.g., \cite{
% SemEval2016:task3:ConvKN,
% DBLP:conf/cikm/TymoshenkoBM16,
% DBLP:conf/cikm/MartinoBR0M16}. 

%In contrast with the previous two editions, this time many more learning methods were used in addition to SVMs, which still yield the best performance overall. 

This year, we saw variety of machine learning approaches, ranging from SVMs to deep learning.

The \emph{KeLP} system, which performed best on Subtask A, 
%is methodologically the same as in 2016~\cite{SemEval2016:task3:KeLP}. KeLP 
was SVM-based and used syntactic tree kernels with relational links between questions and comments, together with some standard text similarity measures linearly combined with the tree kernel. Variants of this approach were successfully used in related research
%\cite{SemEval2016:task3:ConvKN,
\cite{DBLP:conf/cikm/TymoshenkoBM16,DaSanMartino:CIKM:2016}, as well as in last year's \emph{KeLP} system \cite{SemEval2016:task3:KeLP}.

The best performing system on Subtask C, \emph{IIT-UHH}, was also SVM-based, and it used textual, domain-specific, word-embedding and topic-modeling features. The most interesting aspect of this system is their method for dialogue chain identification in the comment threads, which yielded substantial improvements.

% It can be expanded by:
%In particular, they  proposed a novel algorithm based on construction of a user interaction graph to model potential dialogues. This consists of a user graph and a dialogue graph. Initially, the user graph has the question node and the dialogue graph is empty. We add new users to the graphs according to the time- stamp of their occurrence in the thread. For each new comment, we add edges to each previous user and the question in the user graph for the user who commented. Then we pick the maximum outgoing edge from the user who commented, and add that edge in the dialogue graph. Finally, we find the weakly connected components (WCCs) in the dialogue graph and the users in each such WCC are in mutual dialogue. Note that the user graph at the end of each iteration depicts the

The best-performing system on Subtask B was \emph{SimBow}. They used logistic regression on a rich combination of different unsupervised textual similarities, built using a relation matrix based on standard cosine similarity between bag-of-words and other semantic or lexical relations.

This year, we also saw a jump in the popularity of deep learning and neural networks.
For example, the \emph{Beihang-MSRA} system was ranked second with a result very close to that of \emph{KeLP} for Subtask A. They used gradient boosted regression trees, i.e., XgBoost, as a ranking model to combine (\emph{i})~TF$\times$IDF, word sequence overlap, translation probability, (\emph{ii})~three different types of tree kernels, (\emph{iii})~subtask-specific features, e.g., whether a comment is written by the author of the question, the length of a comment or whether a comment contains URLs or email addresses, and (\emph{iv})~neural word embeddings, and the similarity score from Bi-LSTM and 2D matching neural networks.

\emph{LearningToQuestion} achieved the second best result for Subtask B using SVM and Logistic Regression as integrators of rich feature representations, mainly embeddings generated by the following neural networks: (\emph{i})~siamese networks to learn similarity measures using GloVe vectors \cite{Pennington:2014}, (\emph{ii})~bidirectional LSTMs, (\emph{iii})~gated recurrent unit (GRU) used as another network to generate the neural embeddings trained by a siamese network similar to Bi-LSTM, (\emph{iv})~and convolutional neural networks to generate embeddings inside the siamese network.

\noindent The \emph{bunji} system, second on Subtask C, produced features using neural networks that capture the semantic similarities between two sentences as well as comment plausibility. The neural similarity features were extracted using a decomposable attention model \cite{parikh-EtAl:2016:EMNLP2016}, which can model alignment between two sequences of text, allowing the system to identify possibly related regions of a question and of a comment, which then helps it predict whether the comment is relevant with respect to the question.
The model compares each token pair from the question tokens and comment tokens associating them with an attention weight.
Each question-comment pair is mapped to a real-value score using a neural network with shared weights and the prediction loss is calculated list-wise. 
The plausibility features are task-specific, e.g., is the person giving the answer actually trying to answer the question or is s/he making remarks or asking for more information. Other features are the presence keywords such as \emph{what}, \emph{which}, \emph{who}, \emph{where} within the question. There are also features about the question and the comment length. All these features were merged in a CRF.

%In contrast the systme winning
%KMV: this was a fragment just sitting here. I assume it can be deleted. --> yes thanks

Another interesting system is that of \emph{Talla}, which consists of an ensemble of syntactic, semantic, and IR-based features, i.e., semantic word alignment, term frequency Kullback-Leibler divergence, and tree kernels. These were integrated in a pairwise-preference learning handled with a random forest classifier with 2,000 weak estimators. This system achieved very good performance on Subtask B.

Regarding Arabic, \emph{GW\_QA}, the best-performing system for Subtask D, used features based on latent semantic models, namely, weighted textual matrix factorization models (WTMF), as well as a set of lexical features based on string lengths and surface-level matching. 
WTMF builds a latent model, which is appropriate for semantic profiling of a short text. Its main goal is to address the sparseness of short texts using both observed and missing words to explicitly capture what the text is and is not about. The missing words are defined as those of the entire training data vocabulary minus those of the target document.
The model was trained on text data from the Arabic Gigaword as well as on Arabic data that we provided in the task website, as part of the task. For Arabic text processing, the MADAMIRA toolkit was used.
%\emph{GW\_QA} was ranked first out of three submissions, with a MAP score of 61.16\%, which is an important result as it is the only one outperforming the strong Google rank baseline.

\noindent The second-best team for Arabic, \emph{QU-BIGIR}, used SVM-rank with two similarity feature sets. The first set captured similarity between pairs of text, i.e., synonym overlap, language model score, cosine similarity, Jaccard similarity, etc. The second set used word2vec to build average word embedding and covariance word embedding similarity to build the text representation.

The third-best team for Arabic, \emph{UPC-USMBA}, combined several classifiers, including (\emph{i})~lexical string similarities in vector representations, and (\emph{ii})~rule-based features. A core component of their approach was the use of medical terminology covering both Arabic and English terms, which was organized into the following three categories: body parts, drugs, and diseases. In particular, they translated the Arabic dataset into English using the Google Translate service. The linguistic processing was carried out with Stanford CoreNLP for English and MADAMIRA for Arabic. Finally, WordNet synsets both for Arabic and English were added to the representation without performing word sense disambiguation. 

%Both \emph{QU-BIGIR} and \emph{UPC-USMBA} participated in the Arabic subtask last year, and both of them greatly improved their previous system.
%UPC-USMBA achieved the third position but we note that they and QU-BIGIR team were able to greatly improve their last year system.

%KMV: Should there be a separate conclusion section? --> yep

\section{Conclusions}
\label{sec:conclusion}

We have described SemEval-2017 Task 3 on Community Question Answering, which extended the four subtasks at SemEval-2016 Task 3 \cite{nakov-EtAl:2016:SemEval} with a new subtask on multi-domain question duplicate detection.
Overall, the task attracted 23 teams, which submitted 85 runs; this is comparable to 2016, when 18 teams submitted 95 runs. The participants built on the lessons learned from the 2016 edition of the task, and further experimented with new features and learning frameworks. The top systems used neural networks with distributed representations or SVMs with syntactic kernels for linguistic analysis. A number of new features have been tried as well.

Apart from the new lessons learned from this year's edition, we believe that the task has another important contribution: the datasets we have created as part of the task, and which we have released for use to the research community, should be useful for follow-up research beyond SemEval.

%Last year, the SemEval CQA subtasks attracted a large number of participants. However, as the task design was new, we felt that participants would benefit from a second run, to develop more customized approaches. Additionally, we proposed a completely new Subtask E, Multi-Domain Duplicate Detection, based on StackExchange data.

%The SemEval challenge of 2017 has seen strong innovation in the technology used for CQA, including  the use of different learning algorithms and the design of more powerful features for measuring text similarities. The winning systems typically use ensembles and neural network components, both in supervised or unsupervised fashion.

Finally, while the new subtask E did not get any submissions, mainly because of the need to work with a large amount of data, 
%which makes experimentation with new models more complex,
we believe that it is about an important problem and that it will attract the interest of many researchers of the field.

% TO be done for the final version

%Participants preferred different kinds of features for different subtasks:
%
%\paragraph{Subtask A.} Similarities between question subject vs. comment, question body vs. comment, and question subject+body vs. comment.
%
%\paragraph{Subtask B.} Similarities between the original and the related question at different levels: subject vs. subject, body vs. body, and subject+body vs. subject+body.
%
%\paragraph{Subtask C.} The same from above, plus the similarities of the original question ???subject, body, and full levels??? with the comments from the thread of the related question.
%
%The similarity scores to be used as features were computed in various ways, e.g., the majority of teams used dot product calculated over word $n$-grams ($n$=1,2,3), character 3-grams, or with TF-IDF weighting.
%Or simply using word overlap, i.e., the number of common words between two texts, often normalized, e.g., by question/comment length. Or overlap in terms of nouns or named entities.
%
%Other important features, which were used by most systems, are related to rank, e.g., rank of the comment in the question thread, or rank of the related question in the list of questions retrieved by the search engine for the original question.
%

%%%---------------- A c k n o w l e d g e m e n t s
\section*{Acknowledgements} 
This research was performed in part by the Arabic Language Technologies (ALT) group at the Qatar Computing Research Institute (QCRI), HBKU, part of Qatar Foundation. It is part of the Interactive sYstems for Answer Search ({\sc Iyas}) project, which is developed in collaboration with MIT-CSAIL. This research received funding in part from the Australian Research Council.

%We would like to thank the anonymous reviewers for their constructive comments, which have helped us improve the paper.

%We would like to thank the anonymous reviewers for their constructive comments.

%%%---------------- R e f e r e n c e s
\bibliographystyle{acl_natbib}
%\bibliography{refs,semeval2017}
\bibliography{refs}

\end{document}